%% file: aaai-main.tex
\documentclass[letterpaper]{article} 
\usepackage[draft]{aaai2026}
\usepackage{times}  
\usepackage{helvet}  
\usepackage{courier}  
\usepackage[hyphens]{url}  
\usepackage{graphicx} 
\usepackage{natbib}  
\usepackage{caption} 
\usepackage{algorithm}
\usepackage{algorithmic}

\usepackage{newfloat}
\usepackage{listings}
\DeclareCaptionStyle{ruled}{labelfont=normalfont,labelsep=colon,strut=off} 
\floatstyle{ruled}
\newfloat{listing}{tb}{lst}{}
\floatname{listing}{Listing}
\usepackage{amsmath}
\usepackage{amsfonts}
\usepackage{xspace}
\usepackage{booktabs}
\usepackage{multirow}
\usepackage{tabularx}
\usepackage{subfigure}
\usepackage{makecell}

\newcommand{\ourmethod}{ODE-Diff\xspace} 

\title{Counterfactual Probabilistic Diffusion  with Expert Models}
\author {
    Wenhao Mu, 
    Zhi Cao,
    Mehmed Uludag,
    Alexander Rodríguez
}
\affiliations {
    University of Michigan\\
    \{muwenhao, zhicao, muludag, alrodri\}@umich.edu
}

\usepackage{bibentry}

\begin{document}

\maketitle

\begin{abstract}
Predicting counterfactual distributions in complex dynamical systems is essential for scientific modeling and decision-making in domains such as public health and medicine. However, existing methods often rely on point estimates or purely data-driven models, which tend to falter under data scarcity. We propose a time series diffusion-based framework that incorporates guidance from imperfect expert models by extracting high-level signals to serve as structured priors for generative modeling. Our method, \ourmethod, bridges mechanistic and data-driven approaches, enabling more reliable and efficient causal inference. We evaluate \ourmethod across semi-synthetic COVID-19 simulations, synthetic pharmacological dynamics, and real-world case studies, demonstrating that it consistently outperforms strong baselines in both point prediction and distributional accuracy.
\end{abstract}

\begin{links}
    \link{Code}{https://github.com/complex-ai-lab/ODE-Diff}
\end{links}

\input{introduction}
\input{relatedwork}
\input{problem}

\input{method}

\input{experiment}
\input{conclusion}

\newpage
\bibliography{ref}


\newpage
\appendix
\input{appendix}


\end{document}

%% file: introduction.tex
\section{Introduction}
Understanding how complex dynamical systems respond to external stimuli or intervention policies is a fundamental challenge across many scientific domains. In public health, for example, controlling the spread of infectious diseases requires navigating the interplay between human behavior, pathogen dynamics, and environmental factors. Given observations of past interventions, decision-makers seek to evaluate their effectiveness to inform strategies for preventing and containing future pandemics~\cite{borchering2023public}. In this context, counterfactual prediction provides a principled framework for reasoning about such effectiveness and for designing improved policies moving forward~\cite{feuerriegel2024causal}.

Much of the existing work in counterfactual inference builds on the potential outcomes (POs) framework, which defines treatment effects at the individual level in terms of unobserved counterfactuals. While the Individual Treatment Effect (ITE), defined as the difference between an individual’s potential outcomes under treatment and control, is conceptually central, it is inherently unobservable. As a result, most practical approaches focus on estimating the Conditional Average Treatment Effect (CATE)—the expected treatment effect given covariates—which serves as a less individualized but estimable approximation~\cite{imbens2004nonparametric,lim2018forecasting,bica2020estimating}. While many methods model only the expected outcome under each intervention, recent work has emphasized estimating the full conditional outcome distribution to better capture uncertainty~\cite{kennedy2023semiparametric,kim2018causal,melnychuk2023normalizing}.

To support distributional counterfactual inference, recent work has explored the use of generative models capable of capturing the full conditional outcome distribution. Among these, diffusion models have emerged as a powerful tool due to their strong generative performance in high-dimensional settings~\cite{sanchez2022diffusion,ma2024diffpo,augustin2022diffusion,chao2024modelingcausalmechanismsdiffusion}.
However, applying them to real-world dynamical systems remains challenging due to the scarcity of observational data. 
For instance, in public health, some intervention policies may have only been implemented in a few regions, making it difficult to generalize their effects to unseen contexts~\cite{ihme2021modeling}. This sparsity hinders the reliability and trustworthiness of counterfactual predictions, particularly in high-stakes domains.

In this paper, we investigate how imperfect expert-defined models can be integrated into generative frameworks for counterfactual prediction. Specifically, we propose a method for incorporating expert models expressed as ordinary differential equations (ODEs), which we refer to as \emph{expert ODEs}, to enhance generalization and data efficiency. Mechanistic models like expert ODEs are often considered the “gold standard” for capturing causal relationships and modeling the effects of interventions~\cite{mooij2013ordinary, scholkopf2022causality}. One example of this is the SIR (Susceptible-Infected-Removed) model in epidemiology, which has a long history of informing public health policy decisions~\cite{hethcote_mathematics_2000}. However, despite encoding valuable domain knowledge, these models may suffer from incomplete or misspecified mechanisms that limit their predictive performance when used in isolation~\cite{holmdahl2020wrong, qian2021integrating, rodriguez2024machine}. This limitation motivates the development of methods that can flexibly combine the structure and interpretability of expert models with the expressiveness of generative learning, enabling more robust counterfactual inference in complex dynamical systems. Our work builds on recent advances in physics-informed and knowledge-guided machine learning~\cite{rodriguez2023einns, yin2021augmenting, qian2020and}, yet to our knowledge, this integration has not been explored in the context of causal inference. With this work, we aim to encourage future research that leverages expert models as a valuable complement to data-driven approaches in causal inference.

Our contributions are summarized as follows:
(1) We propose \ourmethod, one of the earliest counterfactual methods to integrate the representational power and generative flexibility of data-driven models for time series forecasting with domain knowledge from expert ODEs, which imperfectly capture complex dynamical systems. To achieve this, we develop a novel classifier-guidance mechanism that incorporates differential equations.
(2) \ourmethod introduces a hybrid counterfactual predictor that serves as a classifier-free guidance signal to the diffusion model. This module combines a Neural ODE with the expert ODE, aligning the machine learning component with the continuous-time formulation of domain-specific mechanistic models.
(3) We validate the effectiveness of our approach on both synthetic and real-world datasets through extensive experiments, using a diverse set of metrics to evaluate distributional fit and the accuracy of estimated treatment effects.

%% file: relatedwork.tex
\section{Related work}
Our work aims to learn the counterfactual density distribution of time series, for which we draw from previous work in causal inference and diffusion models. 

\paragraph{Causal Inference for POs and CATE.}
In the causal inference literature, many existing methods focus on conditional avarage treatment effect (CATE), evaluating treatment effectiveness by estimating the average outcomes of a population~\cite{ alaa2017bayesian,imbens2004nonparametric, nie2021quasi, zhang2021treatment}. Traditional approaches often rely on parameterized models with limited flexibility, such as structural nested models and marginal structural models~\cite{li2021g, robins1986new, robins1994correcting, robins2000marginal, rubin1978bayesian}. Recently, advancements in machine learning have driven significant progress in this field, with numerous neural network-based methods proposed to capture the complex relationships between outcomes, interventions, and covariates~\cite{berrevoets2021disentangled,  lim2018forecasting, melnychuk2022causal}. 

However, CATE captures conditional mean effects but does not reflect the full uncertainty or distributional changes. Therefore, directly predicting potential outcomes (POs), i.e., counterfactual distribution prediction, offers a more informative approach. Recent efforts have been made to estimate the density distribution of counterfactual outcomes~\cite{kunzel2019metalearners, shalit2017estimating, yoon2018ganite}, such as percentile estimation of cumulative distribution functions~\cite{chernozhukov2013inference, wang2018quantile} and semi-parametric methods~\cite{kennedy2023semiparametric}. However, these methods are designed for explicitly estimating densities and face challenges in scaling to high-dimensional settings. The recent advancements in generative models have further driven progress in this area. Notable attempts include using Generative Adversarial Networks (GANs) and Autoencoders for counterfactual prediction~\cite{yoon2018ganite, sohn2015learning}. Nevertheless, most of these works focus on static counterfactual prediction, where covariate and interventions do not change over time. Furthermore, the majority of existing methods do not explore time series settings or scenarios with limited counterfactual data.


\paragraph{Diffusion Models.}
They are used to learn complex distributions from a given dataset and then sample high-quality data from noise~\cite{ho2020denoising, sohl2015deep, song2019generative}. 
Recent time series diffusion models have become a popular research direction, including tasks like time series imputation~\cite{li2023ts, chen2024quantifying}, forecasting~\cite{wang2023diffusion, lin2024specstg}, and anomaly detection~\cite{nag2023difftad, wyatt2022anoddpm}. 
In parallel, some efforts have applied diffusion models to causal inference tasks, such as counterfactual image generation~\cite{jeanneret2022diffusion, augustin2022diffusion, sanchez2022diffusion}, video generation\cite{reynaud2022d, reynaud2023feature}, answering causal queries~\cite{chao2023interventional, khemakhem2021causal, sanchez2021vaca}, or addressing static counterfactual inference problems~\cite{wu2024counterfactual, ma2024diffpo}. However, there has been relatively little research on counterfactual prediction for time series using diffusion models. Moreover, since diffusion models typically require large datasets for training, these methods have not addressed the challenge of generating counterfactual time series in environments with highly sparse data~\cite{rombach2022high}.

It is worth noting that some recent works have proposed physics-informed diffusion models~\cite{huang2024diffusionpde, shao2024data} which integrate PDEs/ODEs into the generative process. However, these approaches typically assume that the underlying physical model is accurate and seek to enforce exact agreement between the PDE-defined variables and the neural network outputs. In contrast, our approach is designed for settings where the expert model may be incomplete or misspecified. Rather than enforcing strict alignment, we use the expert ODE as a soft guide to improve the robustness and reliability of diffusion models in the presence of imperfect domain knowledge.



%% file: problem.tex
\section{Problem Formulation}
As mentioned earlier, our goal is to achieve counterfactual density distribution prediction for time series in scenarios where the dynamics are partially observable, domain knowledge is partially available, and counterfactual data is limited. To this end, our approach combines expert ODEs derived from domain knowledge, neural ODEs with enhanced dynamic modeling capabilities, and diffusion models capable of generating time series density distributions, leveraging the strengths of each component. Our novelty lies in integrating expert ODEs into counterfactual density estimation—an aspect that prior work has not explored. 

\paragraph{Observational Data:}  
We are given a dataset $\mathcal{D}$, which includes i.i.d multivariate time series covariates $X \in \mathcal{X} \subseteq \mathbb{R}^{T+d_x}$, along with the corresponding causal prediction target $Y \in \mathcal{Y} \subseteq \mathbb{R}^T$, and a treatment $A\in\{0,1\}^T$.
Here, $T$ represents the length of the time series, and $d_x$ denotes the feature dimension of the covariates. For example, in the context of COVID-19 time series data, observed hospitalizations and mobility trends can serve as covariates $X$, the observed mortality rate corresponds to $Y$, and whether a mask-wearing policy is in effect is represented by $A$.  

\paragraph{Expert ODEs:}
Additionally, we are given an expert model mathematically expressed as a system of ODEs to which we refer to as \emph{expert ODEs}. They contain a system of time-evolving expert variables \( \mathbf{z}^e \in \mathbb{R}^E \), representing domain-specific dynamics, governed by an expert model:
\begin{align}
    \label{eq:expert_ode}
    \dot{\mathbf{z}}^e(t) = f^e(\mathbf{z}^e(t), \mathbf{a}(t); \theta^e),
\end{align}
where \( f^e: \mathbb{R}^{E+1} \to \mathbb{R}^E \) defines the evolution of expert variables based on system-specific parameters \( \theta^e \).  
It is important to note that the effect of the treatment \( A \) is directly incorporated into the mechanistic model via \( \mathbf{a}(t) \). 
In our experiments, we use two different expert ODEs which parameterize Equation~\ref{eq:expert_ode}: an epidemiological ODE for COVID-19 with school closure as treatment (SEIRM model~\cite{wu2020nowcasting}) and a pharmacological ODE for viral load with Dexamethasone dose as treatment (PKPD model~\cite{leon2023modelling}). The details for these two expert ODEs can be found in our appendix.

\paragraph{Remark on Expert ODEs:} In real-world settings, domain experts often provide mechanistic models that reflect their best understanding of system dynamics, even if imperfect~\cite{salsa2015partial,metcalf2020mathematical,holmdahl2020wrong}. For example, an epidemiologist or pharmacologist may supply a fully specified system of ODEs (epidemiological or pharmacological) representing the most reliable model they can offer, along with parameter values and initial conditions calibrated to their expert judgment. This expertise is critical, as the parameters $\theta^e$ and initial conditions $\mathbf{z}^e(0)$ are latent (i.e., not directly observable) and must be inferred from partial and often noisy observations. Such settings frequently give rise to ill-posed inverse problems, where multiple solutions may be consistent with the observations~\cite{karniadakis2021physics}; expert knowledge is therefore essential to constrain the solution space and enable more reliable inference. In this work, we assume an interdisciplinary setting in which we have access to expert-specified parameters $\theta^e$ and initial conditions $\mathbf{z}^e(0)$, which define the ODE trajectories under different interventions. 


\paragraph{Potential outcomes:} Our main interest lies in the individualized treatment effects from observational data.
Therefore, we adopt the standard Neyman-Rubin potential outcomes framework~\cite{rubin2005causal}. Given a realization of treatment $A$, denoted as $a$, we denote $Y(a)$ as the potential outcome under $a$. In this setting, each individual has two potential outcomes: $Y(1)$ if the treatment is administered, $Y(0)$ if the treatment is not administered. However, these two potential outcomes cannot be observed simultaneously in reality. Formally: $Y = AY(1) + (1-A)Y(0)$. In this context, to ensure the identifiability of the potential outcome distribution $P(Y|X,A=a)$, we make three causal assumptions (consistency, unconfoundedness, and overlap)~\cite{ma2024diffpo, curth2021nonparametric, kennedy2023towards}, which are detailed in our appendix.

\paragraph{Generation goal:}
Our goal is to generate the time series potential outcome distribution given a specific treatment $a$ with covariate $X$, formally, $p(Y|X,A=a)$. By generating the entire dynamic distribution, we can better capture the uncertainty in the potential outcomes. 

%% file: method.tex
\section{Our Method}


While data-driven diffusion models exhibit strong generative capabilities, they often struggle in settings with limited data or distribution shifts, as demonstrated in our experiments.
In contrast, expert ODEs, though often incomplete or misspecified, encode qualitative trends grounded in domain knowledge. For instance, a domain expert might not be able to predict the precise infection rate under a new intervention, but they may reliably expect that the rate of infection will decrease~\cite{holmdahl2020wrong}. To leverage such insights, we propose \ourmethod, a framework that integrates expert ODEs in two complementary ways:
\begin{enumerate}
    \item We introduce a \emph{hybrid counterfactual predictor} (Hybrid-CP) that models the co-evolution of expert variables $\mathbf{z}^e$, covariates $X$, and target $Y$ in continuous time. This predictor combines a Neural ODE with the expert ODE, aligning with the continuous-time formulation typical of expert knowledge. It provides an informative point estimate of the counterfactual prediction, which serves as a classifier-free guidance input for our downstream diffusion model.
    \item We develop an \emph{knowledge-guided counterfactual diffusion model} that learns distributions of potential outcomes. This model incorporates a novel classifier guidance mechanism driven by the expert ODE and is embedded within a time series diffusion framework that includes causal reweighting to address treatment selection bias.
\end{enumerate}



\begin{figure}
    \centering
    \includegraphics[width=0.99\linewidth]{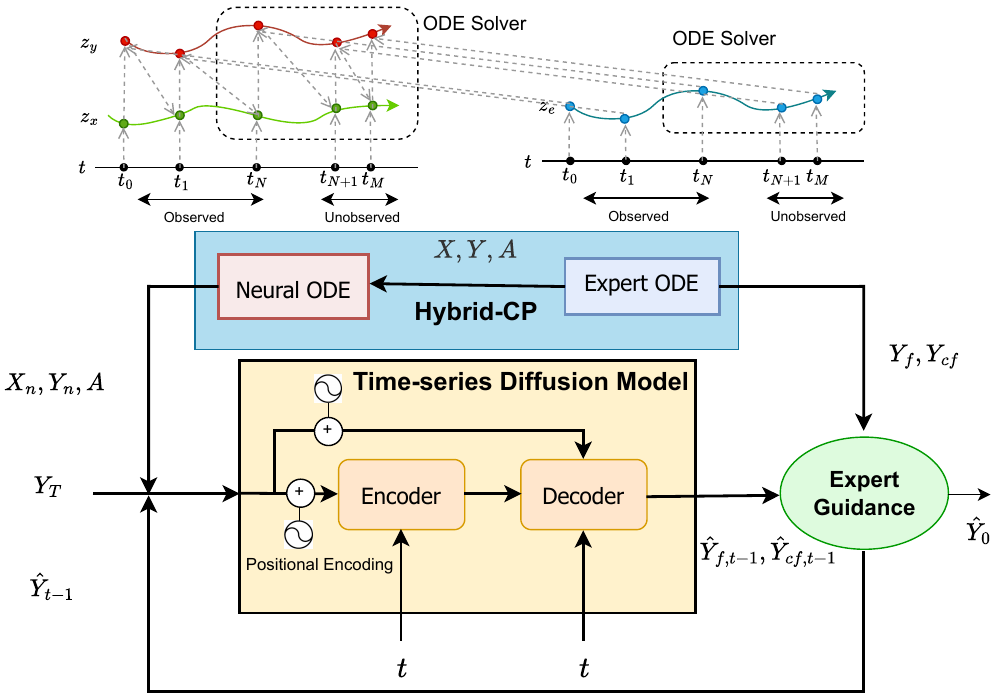}
    \caption{An illustration of \ourmethod. The Expert ODE provides expert variables to the Hybrid-CP and counterfactual information to the guidance of the diffusion model. The Hybrid-CP supplies conditional inputs for diffusion models. The final counterfactual time series is produced by directly generating original time series and iteratively refining them through the guided diffusion process.}
    \label{fig:main}
\end{figure}

\subsection{Hybrid Counterfactual Predictor (Hybrid-CP)}
In practical applications, expert variables \( \mathbf{z}^e(t) \) are not directly observable. Instead, real-world data typically consists of target variables \( \mathbf{y}(t) \), such as symptomatic reports, hospitalization rates, or mortality counts. To bridge this gap, we introduce latent variables \( \mathbf{z}^y(t) \in \mathbb{R}^{M_y} \) that evolve dynamically according to:
\begin{equation}
\begin{aligned}
    \dot{\mathbf{z}}^y(t) &= f^y(\mathbf{z}^y(t), \mathbf{z}^e(t), \mathbf{z}^x(t), \mathbf{a}(t); \theta^y),
\end{aligned}
\end{equation}
where \( f^y : \mathbb{R}^{{M_y} \times E \times {M_x} + 1} \to \mathbb{R}^{M_y} \) is a neural network with (unknown) weights \( \theta^y \).

A key component of our hybrid model is the explicit modeling of covariate dynamics. In counterfactual prediction, many important covariates (e.g., mobility trends, environmental factors, policy enforcement levels) evolve independently but also interact with other latent factors. Rather than treating these covariates as static or exogenously determined, we model them using a separate co-evolving ODE:
\begin{equation*}
\begin{aligned}
    \dot{\mathbf{z}}^x(t) &= f^x(\mathbf{z}^x(t), \mathbf{z}^y(t-1), \mathbf{a}(t); \theta^x),
\end{aligned}
\end{equation*}
where \( f^x : \mathbb{R}^{M_x \times M_y + 1} \to \mathbb{R}^{M_x} \) is a neural network with (unknown) weights \( \theta^x \). Here, the evolution of \( \mathbf{z}^x(t) \) is influenced by the past states of \( \mathbf{z}^y(t) \), which encodes dependencies between the latent measurements and covariates over time. Unlike previous methods that consider covariates as static inputs, our approach explicitly models their temporal evolution, ensuring that counterfactual predictions remain robust even when covariates change dynamically in response to interventions.

The initial states for these variables are determined as follows:
\begin{equation*}
\begin{aligned}
    \mathbf{z}^x(0) &= g_{\xi}(\mathbf{x}(0), \mathbf{a}(0), \mathbf{y}(0)), \\
    \mathbf{z}^y(0) &= g_{\zeta}(\mathbf{z}^x(0), \mathbf{a}(0), \mathbf{y}(0)).
\end{aligned}
\end{equation*}
Similarly, the expert ODE initial state \( \mathbf{z}^e(0) \) is parameterized by a neural network:
\begin{equation*}
\begin{aligned}
    \mathbf{z}^e(0) &= g_{\eta}(\mathbf{x}(0), \mathbf{a}(0), \mathbf{y}(0)),
\end{aligned}
\end{equation*}
where \( g_{\eta} : \mathbb{R}^{d+1+1} \to \mathbb{R}^E \). This ensures that domain knowledge is preserved while allowing adaptability to different initial conditions. In practice, after generating the initial state, a normalization step is typically applied to ensure the data aligns with real-world constraints. For example, in the SEIRM model in epidemiology, we require the sum of all compartments to equal the total population.
Finally, we establish mappings from latent variables to observed measurements:
\begin{equation}
\begin{aligned}
    \mathbf{y}(t) &= g_y(\mathbf{z}^e(t), \mathbf{z}^y(t), \mathbf{z}^x(t), \mathbf{a}(t)), \\
    \mathbf{x}(t) &= g_x(\mathbf{z}^x(t), \mathbf{a}(t)).
\end{aligned}
\end{equation}
where \( g_y : \mathbb{R}^{E \times {M_y} \times {M_x} + 1} \to \mathbb{R} \) and \( g_x : \mathbb{R}^{{M_x} + 1} \to \mathbb{R} \) are neural networks with (unknown) weights \( \gamma_y \) and \( \gamma_x \), respectively, mapping the latent space to the measurement space for \( \mathbf{y}(t) \) and \( \mathbf{x}(t) \). This Hybrid-CP formulation integrates expert ODEs, latent dynamics, and co-evolving covariates, making it well-suited for counterfactual prediction in partially observed systems.

\subsection{Knowledge-guided Counterfactual Diffusion Model}
We design a time series diffusion model for counterfactual prediction that integrates Hybrid-CP forecasts of covariates and the target as classifier-free guidance, and incorporates structural aspects of expert ODEs through a classifier-based guidance mechanism. 

\paragraph{Time Series Diffusion Model.}
Our approach is built upon diffusion models~\cite{song2019generative,ho2020denoising}, particularly those designed for time series~\cite{yuan2024diffusion}. A typical diffusion model is defined by a forward process $q$ and a reverse process $p$. In the forward process, Gaussian noise is gradually added to the data distribution in a Markov chain manner: $q(y_\tau|y_{\tau-1}):=\mathcal{N}(y_\tau;\sqrt{1-\beta_\tau} y_{\tau-1},\beta_\tau \mathbf{I})$ to make it isotropic Gaussian noise in the end of time $T_{d}$, where $\beta_\tau$ is the variance schedule. The reverse process progressively removes the noise to recover the data distribution, and this denoising process $p_\theta(y_{\tau-1}|y_\tau)=\mathcal{N}(y_{\tau-1};\mu_\theta(y_\tau,\tau),\Sigma_\theta(y_\tau,\tau))$ is learned by a neural network. This task of denoising $x_{T_d}$ via the reverse diffusion process can be reformulated as learning a approximator to parameterize $\mu(y_\tau,\tau)$ across all time steps $\tau$. ~\cite{ho2020denoising} proposed to train the approximator $\mu_\theta(y_\tau,\tau)$ using a weighted mean squared error objective which is
\begin{equation}
    L_t = \mathbb{E}_{\tau\sim[1,{T_d}],y_0,\epsilon_\tau}[\| \mu(y_\tau,y_0)-\mu_\theta(y_\tau,\tau)\|^2]
\end{equation}
Here $\mu(x_\tau,x_0)$ denotes the posterior mean of $q(x_\tau|x_0,x_\tau)$. The corresponding training objective can be interpreted as maximizing a weighted variational lower bound of the data log-likelihood. For the details, please refer to appendix.

It is important to note that $\mu_\theta(x_\tau,\tau)$ can be parameterized either in terms of $\epsilon_\theta(y_\tau,\tau)$ or directly in terms of $\hat{y}_0(y_\tau,\tau,\theta)$. Unlike the conventional choice of predicting $\hat{y}_0(y_\tau,\tau,\theta)$, we adopt a direct prediction of $\hat{y}_0(y_t,t,\theta)$ to enhance the performance of time series generation and to simplify the incorporation of Expert ODE guidance. Following ~\cite{yuan2024diffusion}, the training objective can be simplified as:
\begin{equation}
    \mathcal{L_{\text{simple}}}=\mathbb{E}_{\tau\sim[1,{T_d}],y_0,\epsilon_\tau}[w_\tau\| y_0-\hat{y}_0(y_\tau,\tau,\theta)\|^2]
\end{equation}
where $w_\tau=\frac{\lambda \alpha_\tau(1-\bar{\alpha}_\tau)}{\beta_\tau^2}$ and $\lambda$ is a constant. Then the sampling process can be approximated by:
\begin{equation}
\begin{aligned}
    y_{\tau-1} = \frac{\sqrt{\bar{\alpha}_{\tau-1}}\beta_\tau}{1-\bar{\alpha}_\tau}\hat{y_0}(y_\tau,\tau,\theta)&+\frac{\sqrt{\alpha_\tau}(1-\bar{\alpha}_{\tau-1})}{1-\bar{\alpha_\tau}}y_\tau\\ &+ \frac{1-\bar{\alpha}_{\tau-1}}{1-\bar{\alpha}_\tau}\beta_\tau z_\tau
\end{aligned}
\end{equation}
where $z_\tau\sim \mathcal{N}(0,\mathbf{I})$, $\alpha_\tau=1-\beta_\tau$ and $\bar{\alpha}_\tau=\prod_{s=1}^\tau\alpha_s$


\paragraph{Expert ODE Guidance.} 
To efficiently leverage the reliability provided by domain knowledge encoded in the expert ODE, while avoiding the bias introduced by the simplifications inherent in mechanistic models, we design a set of effective guidance strategies. To incorporate this guidance into generation, we use a classifier guidance mechanism to steer the diffusion process toward trajectories that match the qualitative patterns of the expert.

Let $y_\tau$ be the noisy sample at diffusion timestep $\tau$, and let $\hat{y}_0(y_\tau, \tau, \theta)$ ($\hat{y}_0$ in short) denote the model’s counterfactual prediction—i.e., the denoised estimate of the counterfactual trajectory. Let $y_0$ denote the observed factual trajectory from the data. We define the guided prediction as:
\[
    {\Tilde{y}_0}(y_\tau,\tau,\theta) = \hat{{y_0}}(y_\tau,\tau,\theta) + \eta\nabla_{y_\tau}\text{Loss}_{\text{cf}}
\]
Here, $\eta$ is the guidance strength parameter, and the counterfactual loss $\text{Loss}_{\text{cf}}$ is defined as:
\begin{equation}
\begin{aligned}\text{Loss}_{\text{cf}} =& \sum_{t}\|\mathbf{g}(\hat{y}_0,y_0,t)-\mathbf{g}(f_{cf}^e,f_{f}^e,t)\|_2^2 \\ &+ \|\mathbf{g'}(\hat{y}_0,y_0,t)-\mathbf{g'}(f_{cf}^e,f_{f}^e,t)\|_2^2
\end{aligned}
\end{equation}
Here, $\mathbf{g}(y_1,y_2,t)$ and $\mathbf{g'}(y_1,y_2,t)$ are functions used to estimate the value and directional relationships between $y_1$ and $y_2$ at time $t$. This loss penalizes discrepancies between the generated counterfactual relationships and the knowledge-derived guidance, both in value and direction. The result is a model that produces counterfactual trajectories that are qualitatively consistent with domain knowledge, while preserving the generative flexibility of diffusion-based models.

A key bottleneck in classifier-guided diffusion models for counterfactual prediction is determining the proper guidance strength $\eta$. To address this and fully leverage domain knowledge embedded in the expert ODE, we design a heuristic two-step procedure as shown in Algorithm 1. First, we align the factual simulation of the expert ODEs with the observed factual data and apply the same alignment scheme to the counterfactual simulation. Then, we compute the correlation between the counterfactual prediction of the diffusion models and that of the expert ODEs under different values of $\eta$, and select the $\eta$ yielding the highest correlation. This method effectively utilizes domain knowledge while minimizing errors introduced by the rigidity of mechanistic models.

\begin{algorithm}
\caption{Heuristic Selection of Guidance Strength $\eta$ in Knowledge-Guided Counterfactual Diffusion}
\begin{algorithmic}[1]
\REQUIRE Observed factual data $y_\text{f}$, factual expert ODE simulation $y^\text{ODE}_\text{f}$, counterfactual expert ODE simulation $y^\text{ODE}_\text{cf}$, counterfactual diffusion model outputs $y^\text{Diff}_\text{cf}$, set of candidate guidance strengths $\{\eta_i\}$

\STATE \textbf{Step 1: Factual Alignment}
\STATE Align expert ODE's factual simulation $y^\text{ODE}_\text{f}$ with observed data $y_\text{factual}$ using dynamic alignment (e.g., DTW or peak matching)
\STATE Apply the same alignment transformation to expert ODE's counterfactual simulation $y^\text{ODE}_\text{cf}$ to get $\tilde{y}^\text{ODE}_\text{cf}$

\STATE \textbf{Step 2: Correlation-Based Selection}
\FOR{each candidate $\eta_i$ in $\{\eta_i\}$}
    \STATE Run diffusion model with classifier-guidance strength $\eta_i$ to generate $y^\text{Diff}_\text{cf}(\eta_i)$
    \STATE Compute correlation $r_i = \text{Corr}(y^\text{Diff}_\text{cf}(\eta_i), \tilde{y}^\text{ODE}_\text{cf})$
\ENDFOR

\STATE Select $\eta^* = \arg\max_{\eta_i} r_i$
\RETURN $\eta^*$
\end{algorithmic}
\end{algorithm}

\paragraph{Factual guidance.}
We also introduce a factual loss, which
is derived from the consistency between the counterfactual and factual data observed prior to the change in treatment.
Details about factual loss can be found in appendix.


\paragraph{Causal Reweight Score.}
We adopt the counterfactual diffusion loss~\cite{wu2024counterfactual} inspired by inverse probability of treatment weighting (IPTW), which enables unbiased learning with potentially biased observational data. The key challenge in counterfactual learning is the distribution shift due to selection bias, which requires reweighting observational data using the estimated propensity score $\hat{\pi}(x_t,a_{t-1}) = g_{\phi}(x_t,a_t)$. The weighting function is defined as:
\[
w_{\hat{\pi}}(x, a) = \frac{1}{\prod_{t-d+1}^T\hat{\pi}(x_t,a_{t-1})}
\]
where $d$ is the history dependence length. 
To integrate the time series diffusion model with point estimation prior $y'$ and covariate $x$, we employ a conditional loss objective during training that incorporates both time series forecasting and a reweighted score function. It can realize a efficiently and unbiasedly training for the diffusion model to learn the causal relationship. The training objective is formally written as:
\begin{equation}
\begin{aligned}
    \mathcal{L}(\theta) = \mathbb{E}_{\tau\sim[1,T_d],\; (y_0, x, a) \sim p(Y, X, A),\, \epsilon_\tau}
    \big[ w_{\hat{\pi}} w_\tau \big\| y_0 - {} \\
    \hat{y}_0(y_\tau,\tau,\theta \mid y',x,a) \big\| \big]
\end{aligned}
\end{equation}
where $w_\pi$ is the reweight score based on the propensity score and $a$ is the treatment.

%% file: experiment.tex
\section{Experimental Setup}

\subsection{Baselines and Metrics}
\noindent\textbf{Baselines.}
We include state-of-the-art models for counterfactual distribution prediction, and since our approach is built on neural ODEs, we also compare against a prominent neural ODE-based counterfactual method: (1) MS-Diffusion~\cite{wu2024counterfactual}, a generative model that learns counterfactual distributions using the IPTW (Inverse Probability of Treatment Weighting) method. (2) DiffPO~\cite{ma2024diffpo}, a diffusion model that learns potential outcome distributions using an orthogonal diffusion loss. (3) GANITE~\cite{shi2019ganite}, a GAN-based counterfactual generative model. (4) TE-CDE~\cite{seedat2022continuous}, a counterfactual neural ODE point estimator to which we incorporate a dropout layer to generate counterfactual distributions. 
The last two are not strictly baselines but standalone components of our final model:
(5) Expert ODE, a domain-specific ODE incorporating partial expert knowledge~\cite{wu2020nowcasting,leon2023modelling}. 
(6) Hybrid Counterfactual Predictor (Hybrid-CP), our neural ODE-based approach that incorporates expert variables from the expert ODE and models their co-evolution with covariates and the target variable.

\noindent\textbf{Metrics.} 
We adopt evaluation metrics from prior work on counterfactual diffusion models~\cite{ma2024diffpo, wu2024counterfactual}—including Wasserstein distance, root mean squared error (RMSE), prediction interval coverage (confidence levels of 75\%, 90\% and 95\%), and Conditional Average Treatment Effect (CATE) error—and extend them with additional measures for uncertainty quantification (calibration score) and alignment with temporal trends (Pearson correlation). We briefly describe Wasserstein distance, CATE, and calibration score, and refer the reader to our appendix for more details on all of them.

The \(k\)-Wasserstein distance measures the discrepancy between two probability distributions; in our experiments, we report results using \(k = 1\), where lower values indicate better alignment. Since our method directly generates the potential outcome distribution, we can use it to compute CATE. Specifically, we first estimate the mean of the generated potential outcomes, \(\mathbb{E}[Y \mid X, A = a]\) for all \(a\), and then compute CATE using its definition:
$\mathbf{CATE}(x) = \mathbb{E}[Y \mid X, 1] - \mathbb{E}[Y \mid X, 0]$.
We evaluate the quality of generated CATE values against ground truth using RMSE, with lower scores indicating better performance. The calibration score~\cite{kamarthi2022camul,xu2021conformal} assesses how well the predicted intervals reflect empirical uncertainty across multiple confidence levels; following prior work~\cite{li2025neural}, we evaluate this across 11 levels.

\subsection{Datasets}
We evaluate \ourmethod on one fully synthetic dataset and one semi-synthetic dataset, both of which provide access to factual and counterfactual outcomes. This allows for rigorous benchmarking as we have the ground-truth counterfactual distribution. In addition, we include a real-world dataset, which we use for a case study.

\begin{table*}[ht]
\centering
\caption{Comparison of our proposed \ourmethod model with baseline models on our synthetic datasets. The table evaluates performance using multiple metrics, including Wasserstein distance, RMSE, prediction intervals (75\%, 90\%, 95\%), CATE, calibration score, and correlation. We bold the best method and underline the second best.}
\resizebox{\textwidth}{!}{
\begin{tabular}{lccccccccc}
\toprule
\multicolumn{9}{c}{\textbf{COVID-19}}\\
\toprule
\textbf{Model} & \textbf{Wasserstein distance} & \textbf{RMSE} & \textbf{75\% PI} & \textbf{90\% PI} & \textbf{95\% PI} & \textbf{CATE} & \textbf{Calibration Score} & \textbf{Corr}   \\ 
\midrule
MS-Diffusion  & $0.2651\pm0.0015$   & $0.3639\pm0.0038$   & $\underline{0.1068\pm0.0559}$   & $0.2259\pm0.0364$  & $0.3185\pm0.0279$   & $0.3578\pm0.0068$ & $0.4509\pm0.0349$ & $0.9253\pm0.0008$ \\
DiffPO  & $0.2058\pm0.0104$   &  $0.2776\pm0.0147$  &  $\mathbf{0.1053\pm0.0211}$  &  $0.1444\pm0.0226$  & $0.1934\pm0.0082$  & $0.2795\pm0.0130$ & $0.4859\pm0.0104$ & $0.9802\pm0.0018$ \\
GANITE & $4.3618\pm0.1696$ & $4.5443\pm0.1757$ & $0.0000\pm0.0000$ & $0.0000\pm0.0000$ & $0.0000\pm0.0000$ & $0.7164\pm0.0000$ & $0.5845\pm0.0000$ & $0.2816\pm0.0350$ \\
TE-CDE & $0.4667\pm0.0295$ & $0.5848\pm0.0817$ & $0.1755\pm0.1055$ & $\mathbf{0.2462\pm0.1392}$ & $\underline{0.2892\pm0.1498}$ & $0.5648\pm0.0799$ & $\underline{0.4386\pm0.0802}$ & $0.4215\pm0.0540$ \\
\midrule
Hybrid-CP  &  $\underline{0.0839\pm0.0104}$  &  $\underline{0.1145\pm0.0132}$  &  -  &  -  & -  & $\underline{0.1095\pm0.0005}$ & - & $\underline{0.9818\pm0.0222}$\\
Expert-ODE  &  $0.4446\pm0.0000$  &  $0.5626\pm0.0001$  &   - & -   & - & $0.2123\pm0.0001$ & - & $0.9809\pm0.0003$ \\
\midrule
\ourmethod & $\mathbf{0.0506\pm0.0066}$ & $\mathbf{0.0613\pm0.0075}$ & $0.1470\pm0.0647$ & $\underline{0.2459\pm0.0453}$ & $\mathbf{0.4209\pm0.0724}$ &
$\mathbf{0.0612\pm0.0085}$ & $\mathbf{0.4135\pm0.0173}$ & $\mathbf{0.9832\pm0.0057}$ \\

\bottomrule
\end{tabular}}

\resizebox{\textwidth}{!}{
\begin{tabular}{lcccccccc}
\multicolumn{9}{c}{\textbf{Dexamethasone}}\\
\toprule
\textbf{Model} & \textbf{Wasserstein distance} & \textbf{RMSE} & \textbf{75\% PI} & \textbf{90\% PI}& \textbf{95\% PI} & \textbf{CATE} & \textbf{Calibration Score} & \textbf{Corr}\\ 
\midrule
MS-Diffusion  &  $\underline{0.2094\pm0.0026}$ &  $\underline{0.2608\pm0.0028}$  &  $\mathbf{0.5768\pm0.0078}$  &  $\mathbf{0.7537\pm0.0128}$  &  $\mathbf{0.8182\pm0.0127}$  & $\underline{0.1894\pm0.0010}$ & $\mathbf{0.1556\pm0.0025}$ & $0.6861\pm0.0069$  \\
DiffPO & $0.2162\pm0.0373$ &  $0.3076\pm0.0464$  &  $0.2364\pm0.0131$  &  $0.3440\pm0.0364$  &  $0.4177\pm0.0534$  &  $0.2881\pm0.0382$ & $0.3763\pm0.0148$ & $\underline{0.7031\pm0.0992}$  \\
GANITE  &  $0.8410\pm0.0044$  &  $0.9430\pm0.0092$  &   $0.0000\pm0.0000$ &  $0.0000\pm0.0000$  & $0.0000\pm0.0000$  &  $0.8357\pm0.0138$ & $0.5845\pm0.0000$ & $0.3589\pm0.8954$  \\
TE-CDE  & $0.3553\pm0.0078$   &  $0.4804\pm0.0084$  & $0.0866\pm0.0093$   &  $0.1217\pm0.0069$  &  $0.1422\pm0.0111$  & $0.3164\pm0.0027$ & $0.5109\pm0.0076$ & $0.6515\pm0.0355$ \\
\midrule
Hybrid-CP  & $0.3056\pm0.0056$  & $0.4029\pm0.0079$ & - & - & - & $0.8020\pm0.0203$ & - & $0.6453\pm0.0084$ \\
Expert-ODE  &  $0.3851\pm0.0006$  &  $0.4829\pm0.0005$  &  -  &  -  & - & $0.1730\pm0.0001$ & - & $0.3453\pm0.0014$ \\
\midrule
\ourmethod & $\mathbf{0.1500\pm0.0041}$ & $\mathbf{0.1930\pm0.0041}$ & $\underline{0.4240\pm0.0144}$ & $\underline{0.5888\pm0.0295}$ & $\underline{0.6755\pm0.0024}$ &
$0.2058\pm0.0126$ & $\underline{0.2350\pm0.0121}$ & $\mathbf{0.8751\pm0.0024}$ \\

\bottomrule
\end{tabular}}
\end{table*}

\subsubsection{Semi-synthetic COVID-19 Data.} 
We constructed a semi-synthetic dataset simulating the impact of mask mandates across 121 metropolitan areas in the United States during the COVID-19 pandemic. Ground truth data were generated using the SEIR-HD epidemiological model~\cite{kain2021chopping}, covering a 52-week period from 2020 to 2021. The treatment variable is the mask mandate policy (encoded as 0 for no mandate and 1 for mandate), while covariates include new hospitalizations and symptomatic infectious cases. The outcome variable is the number of COVID-19-related deaths per 1,000 people at the city level. Cities were grouped into “strict” and “relaxed” policy categories based on the comprehensiveness and duration of their mandates~\cite{nguyen2021impact}, with most strict-policy cities implementing mandates around week 15~\cite{chernozhukov2021causal}. Initial conditions were derived from U.S. Census data~\cite{uscensus2020} and epidemiological estimates of infectious-to-exposed ratios were taken from relevant literature~\cite{gandhi2021first}. To capture real-world variability, the simulation incorporates heterogeneity in hospitalization rates and transmission dynamics across cities.

\subsubsection{Fully-synthetic Dexamethasone Data.}  
We simulate a fully synthetic dataset using a pharmacological model adapted from~\cite{dai2021prototype}, which describes the evolution of five physiological variables under dexamethasone treatment in COVID-19 patients. The simulation includes 50 patients over a time horizon of $T = 14$ days, aligning with the median length of hospital stay for COVID-19 patients. The treatment variable is whether an individual receives the treatment,
for which we use a scaling parameter to modulate the variation in dose levels~\cite{qian2021integrating}. To reflect the real-world sparsity of treatment timing data, we fix the treatment to occur at $t = 3$, based on empirical observations that dexamethasone is typically administered around the third day of hospitalization~\cite{Horby2021Dexamethasone}. For each patient, the initial condition is randomly generated, with each component independently drawn from an exponential distribution with rate $\lambda = 100$. In the training set, half of the patients are randomly assigned to receive a one-time treatment, while the remaining patients receive no treatment. 

\subsubsection{COVID-19 Policy Real-world Dataset}
Policy decisions such as school closures can have immediate and far-reaching impacts on epidemic outcomes. While simulated experiments offer controlled settings to test models, real-world data is useful for demonstrating the practical utility of counterfactual learning in public health decision-making. To assess our model’s real-world performance, we evaluate its predictions against baseline methods in scenarios where accurate forecasting of outcomes—such as hospitalizations and deaths—following policy interventions is critical. We focus on the Delta variant surge in the U.S., using data from all states during weeks 27 to 45 of 2021. In alignment with our simulated experiments, we use weekly hospitalization counts as covariates, weekly death counts as the outcome, and a binary treatment indicator for school closure policies. The treatment is defined as \texttt{0} for weak closure policies and \texttt{1} for stricter policies. The dataset is sourced from the Oxford COVID-19 Government Response Tracker\footnote{\url{https://github.com/OxCGRT/covid-policy-dataset}}. 

\subsection{Expert ODEs}
\ourmethod leverages expert ODEs to enable more robust and causally meaningful prediction of dynamics. For the COVID-19 experiments, we used the SEIRM model~\cite{wu2020nowcasting}, and for the Dexamethasone drug response experiments, we used the PKPD model~\cite{leon2023modelling}. It is important to emphasize that the expert models used in our framework are simpler than those used to generate the synthetic data. For instance, the SEIRM model lacks a hospitalization compartment, and the adaptive immunity component does not exist in the PKPD model.
Detailed comparison can be found in appendix. 

\begin{table*}[!t]
\centering
\caption{Ablation of \ourmethod on synthetic Dexamethasone data. We bold the best method and underline the second best.}
\resizebox{\textwidth}{!}{
\begin{tabular}{lccccccccc}

\toprule
\textbf{Model} & \textbf{WD}  & \textbf{RMSE} & \textbf{75\% PI} & \textbf{90\% PI} & \textbf{95\% PI} & \textbf{CATE}  & \textbf{Calibration Score} & \textbf{Corr} \\ 
\midrule

w/o Hybrid ODE & $0.2162\pm0.0373$ &  $0.3076\pm0.0464$  &  $0.2364\pm0.0131$  &  $0.3440\pm0.0364$  &  $0.4177\pm0.0534$  &  $0.2881\pm0.0382$ & $0.3763\pm0.0148$ & $0.7031\pm0.0992$  \\
w/o guidance & $0.1515\pm0.0007$ & $0.2329\pm0.0048$ & $0.3573\pm0.0232$ & $0.4973\pm0.0288$ & $0.5724\pm0.0405$ & $0.2119\pm0.0054$ & $0.2790\pm0.0190$ & $0.8442\pm0.0048$ \\
w/o value guidance & $0.1545\pm0.0020$ & $0.2052\pm0.0050$ & $\underline{0.3724\pm0.0115}$ & $\underline{0.5453\pm0.0085}$ & $\underline{0.6240\pm0.0268}$ & $\underline{0.1954\pm0.0034}$ & $\underline{0.2633\pm0.0063}$ & $\underline{0.8681\pm0.0069}$ \\
w/o direction guidance & $\mathbf{0.0752\pm0.0037}$ & $\mathbf{0.1309\pm0.0058}$ & $0.1088\pm0.0081$ & $0.1813\pm0.0154$ & $0.2324\pm0.0266$ & $\mathbf{0.1502\pm0.0071}$ & $0.4765\pm0.0089$ & $0.9394\pm0.0102$ &\\
\midrule
\ourmethod & $\underline{0.1500\pm0.0041}$ & $\underline{0.1930\pm0.0041}$ & $\mathbf{0.4240\pm0.0144}$ & $\mathbf{0.5888\pm0.0295}$ & $\mathbf{0.6755\pm0.0024}$ &
$0.2058\pm0.0126$ & $\mathbf{0.2350\pm0.0121}$ & $\underline{0.8751\pm0.0024}$ \\

\bottomrule
\label{tab:ablation}
\end{tabular}}
\end{table*}

\section{Results}
In this section, we evaluate our approach on our high-fidelity synthetic datasets, present a case study using real-world data, and perform ablation studies to assess the contributions of our proposed method.
\subsection{Performance in Synthetic Data}

Table 1 presents the results on the semi-synthetic COVID-19 dataset and the synthetic Dexamethasone dataset. We observe that, compared to advanced generative models and neural ODEs, our method achieves state-of-the-art performance on both datasets and obtains the highest average ranking. The COVID-19 dataset poses a significant learning challenge, and most methods struggle to capture an accurate counterfactual distribution. In contrast, our approach demonstrates a clear improvement. On the Dexamethasone dataset, the standalone expert ODE does not perform strongly, but when combined with our proposed guidance strategy, it effectively empowers diffusion models to outperform other methods in terms of both accuracy and distribution shape.

It is also worth noting that the Hybrid-CP achieves the best performance among Neural ODE-based models on both datasets, which can be attributed to the prior knowledge provided by the expert variables. The overall low PI scores in the COVID-19 experiments are due to the long sequence lengths and significant individual-level variability, which make it particularly challenging to accurately predict the direction and shape of counterfactual trajectories. Moreover, since Hybrid-CP and Expert ODE are deterministic mechanistic models and are integrated components of our generative framework, we do not treat them as strict baselines, nor do we introduce additional sources of uncertainty—therefore, no PIs are reported for them.


\subsection{Case Study: Predicting Outcomes Post Policy Change in Real-world Data}
Evaluating the causal effect of interventions such as school closures on public health outcomes is challenging because we cannot directly observe the counterfactual scenario. We address this by constructing a data-driven \emph{proxy} for the counterfactual policy impact. Specifically, for each region we identify similar regions with Dynamic Time Warping (DTW) on the pre-intervention death trajectories. We then label regions as ``strong'' or ``weak'' policy adopters based on whether they implement school closures in the upcoming period. By computing the Wasserstein distance between the distributions of post-period outcomes in these two sets, we create a surrogate measure of the policy effect. Our \ourmethod is then used to predict counterfactual outcomes under flipped policies. We train the model on factual trajectories (including the observed policy sequence) and then simulate forward by altering the school-closure input for each test region. In other words, we ask: what would the death trajectory have been if a region that closed schools had kept them open (and vice versa)? This yields a model-based estimate of the outcome difference between strong vs.\ weak policy groups. 

\begin{figure}
    \centering
    \includegraphics[width=0.8\linewidth]{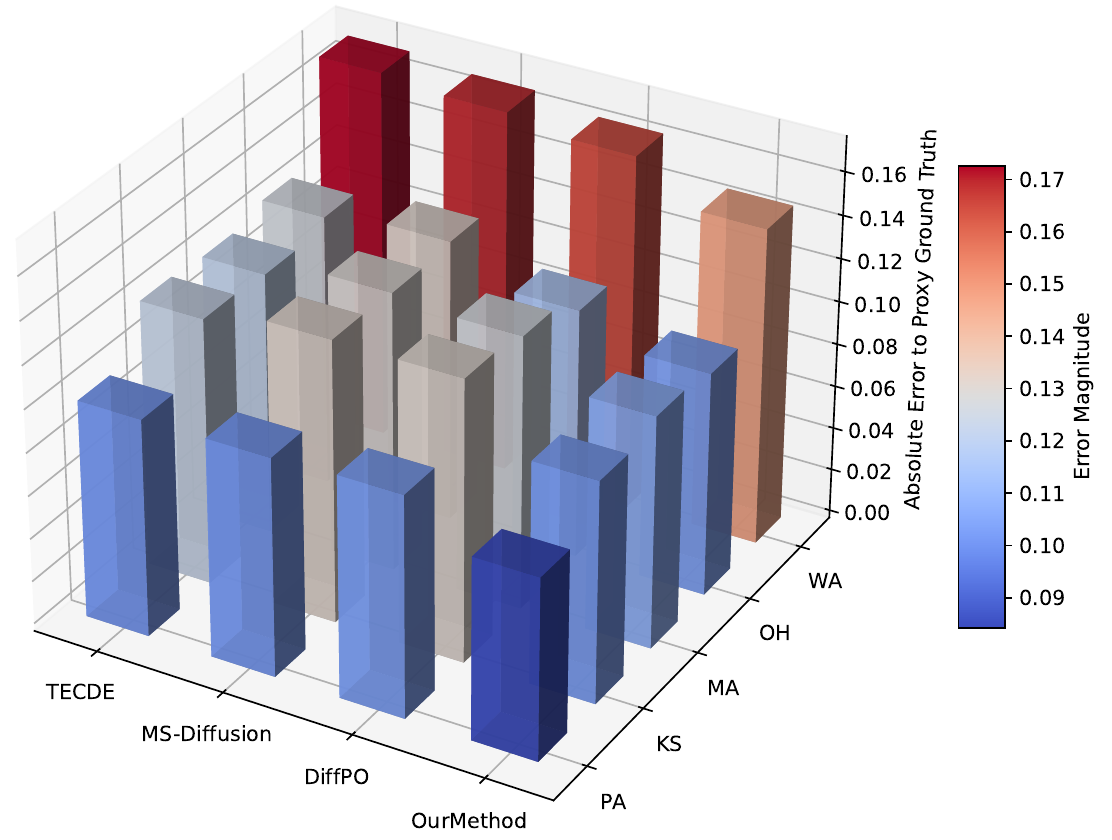}
    \caption{The gap between counterfactual predictions made by different models and the proxy ground truth values on the real-world dataset.}
    \label{fig:realworld}
\end{figure}

We find that \ourmethod's predictions closely match the data-driven proxy
and outperforms baselines on most states, demonstrating strong model fidelity. We can see a subset of the states in Figure~\ref{fig:realworld}, and the complete results can be found in our appendix.


\subsection{Additional Results}

\noindent\textbf{Ablation study:} We conducted an ablation study to evaluate the impact of different components of our generative framework on overall performance. In Table~\ref{tab:ablation}, we observe that under data-scarce conditions in the Dexamethasone dataset, the Hybrid-CP notably improves performance. Moreover, both the numerical guidance and derivative-based guidance demonstrate strong robustness, and their combination yields the model’s best overall SOTA performance.

\noindent\textbf{Selection of Guidance Strength:}
As noted earlier, our approach does not blindly follow the expert ODE but instead uses it as soft guidance. To support this, we introduced a heuristic selection procedure. In the appendix, we present performance curves that illustrate how our evaluation metrics vary with guidance strength.

%% file: conclusion.tex
\section{Conclusion}
In this work, we proposed a framework for integrating imperfect yet useful causal knowledge of expert ODE models into time series probabilistic diffusion models.
Through this integration, we enabled more robust modeling of counterfactual distributions in complex dynamical systems. To our knowledge, this is the first work to introduce knowledge-guided machine learning into the context of causal inference. Our results demonstrate the effectiveness of this approach and highlight its potential for broader application in other causal inference tasks.


\noindent\textbf{Future outlook.} We see promising opportunities to extend counterfactual knowledge-derived guidance beyond time series modeling. In particular, applications in robotics offer rich potential, as expert knowledge is often available in structured but incomplete form. More broadly, our ideas may apply to other generative tasks--such as image or graph generation--where high-level structural or directional priors exist but detailed quantitative supervision is unavailable. We hope this work encourages further research at the intersection of physics-informed/knowledge-guided machine learning, causal modeling, and generative learning. 


\section*{Acknowledgements}
This work was supported in part by the Centers for Disease Control and Prevention Award NU38FT000002 and start-up funds from the University of Michigan.

%% file: appendix.tex
\section{Extended Related Work}

\noindent\textbf{Neural ODEs.}
Neural ODEs~\cite{chen2018neural} provide a continuous alternative to conventional discrete-layered neural networks. Instead of applying a fixed number of layers, Neural ODEs model the transformation of hidden states as a continuous process defined by an ODE, making them effective for time series and generative models. In time series applications, they have demonstrated promising results in handling irregularly sampled and noisy data~\cite{kidger2020neural,li2020scalable} and capturing complex, co-evolving dynamics~\cite{dupont2019augmented,jhin2021ace}.

A few studies have integrated expert-designed ODEs with neural ODEs to comprehensively describe system dynamics~\cite{qian2021integrating,yin2021augmenting}. However, when applied to counterfactual learning, neural ODE-based methods remain conventional time series models and do not inherently account for challenges such as time-dependent confounding. Extensions like \cite{seedat2022continuous} employ domain adversarial training to learn representations that adjust for time-dependent confounding, making them suitable for causal estimation. Nevertheless, these approaches focus on point estimation of trajectories and fail to capture the full counterfactual distribution, overlooking population heterogeneity. To address this limitation, our work integrates diffusion models to generate the full density distribution of counterfactual time series.

\section{Extended Problem Setting}
\label{sec:causalassumptions}
\noindent\textbf{Causal assumptions:}
(1) Consistency: if a unit receives treatment $A$, then the observed outcome $Y$ is equal to the potential outcome under that treatment condition, $Y(A)$. In other words, the observed outcome corresponds to the potential outcome for the treatment actually received. Formally: $Y=Y(A)$ (2) Ignorability: conditional on covariates X, the treatment assignment $A$ is independent of the potential outcomes $Y(1)$ and $Y(0)$. In other words, after accounting for observed covariates, there are no unmeasured confounders affecting both the treatment assignment and the outcome. Formally: $Y(0),Y(1) \perp\!\!\!\perp A|X$. (3): Positivity: For every possible value of the covariates X, there is a positive probability of receiving either treatment or control. This ensures that there are enough treated and untreated units across all levels of X to allow for causal effect estimation. Formally: $0<P(T=1|x)<1,\forall X$.

\section{Examples of Expert ODE Models}
\label{sec:expertODEs}
In this paper, we use two different expert ODEs which parameterize Eq.~\ref{eq:expert_ode}: an epidemiological ODE and a pharmacological ODE. We introduce them below.

$\bullet$ \textbf{The SEIRM epidemiological model.}  
It represents disease progression through Susceptible, Exposed, Infected, Recovered, and Mortality ($s$,$e$,$i$,$r$,$m$), and has been widely used to model COVID-19 due to its prolonged incubation period~\cite{wu2020nowcasting}. The evolution of these compartments follows:
\begin{gather*}
\label{eq:seirm}
\dot{s} = - \beta s i / N \quad \dot{e}  = \beta s i / N - \alpha e \\\nonumber 
\dot{i} = \alpha e  - \gamma i - \mu i \qquad \dot{r} = \gamma i \qquad \dot{m} = \mu i,
\end{gather*}
where $N$ is the population size and $\beta, \alpha, \gamma, \mu$ are the parameters $\theta^e$. Typically, only mortality is observed, while the other variables remain latent, representing the underlying unobservable epidemic dynamics. We simulate the effect of treatment \( A \) via the decay of \( \beta \), as public policies such as mask mandates restrict population movement and consequently reduce the average contact rate \cite{arik2021ai}.

\
$\bullet$\textbf{The PKPD pharmacological model.} 
The second expert ODE model describes the pharmacokinetics and pharmacodynamics (PKPD) of dexamethasone in regulating immune response and viral replication \cite{qian2021integrating,leon2023modelling}. This model captures the interaction between drug concentration and immune system dynamics, as well as the clearance of dexamethasone. The ODEs governing the PKPD model are defined as follows:
\begin{equation*}
\begin{aligned}
\dot{z}_1 =& k_{IR} \cdot z_4 + k_{PF} \cdot z_4 \cdot z_1 - k_O \cdot z_1 \\ &
+ \frac{E_{\max} \cdot z_1^{h_P}}{\text{EC}_{50}^{h_P}  + z_1^{h_P}}  - k_{\text{Dex}} \cdot z_1 \cdot z_2 
\end{aligned}
\end{equation*}
\begin{gather*}
\dot{z}_2 = -k_2 \cdot z_2 + k_3 \cdot z_3, \qquad \dot{z}_3 = -k_3 \cdot z_3\\
\dot{z}_4 = k_{DP} \cdot z_4 - k_{IIR} \cdot z_4 \cdot z_1 - k_{DC} \cdot z_4,
\end{gather*}
where \( z_1 \) represents the innate immune response, \( z_2 \) and \( z_3 \) describe the concentration of dexamethasone in lung tissue and plasma, and \( z_4 \) models viral replication. The treatment effect is incorporated through \( z_3 \)\cite{qian2021integrating}, which follows a pharmacokinetics-driven decay model, adjusting drug dosage through a scaling parameter \( k_d \). 
A more detailed explanation of this system and the expert ODE for Dexamethasone experiment is provided in the following section.

\section{The pharmacological model for dexamethasone} 
\textit{ODE for Synthetic Dataset (\cite{qian2021integrating})}
The original PKPD ODE model describes the immune system's response to viral infections, incorporating both innate and adaptive immune interactions. The state variables include \( z_1 \), representing the innate immune response (measured by Type I IFNs); \( z_2 \) and \( z_3 \), which describe the dexamethasone concentration in lung tissue and plasma, respectively; \( z_4 \), modeling viral load; and \( z_5 \), representing the adaptive immune response (Cytotoxic T cells). The dynamics of these variables follow:

\begin{equation*}
\begin{aligned}
\dot{z}_1 = & k_{IR} \cdot z_4 + k_{PF} \cdot z_4 \cdot z_1 - k_O \cdot z_1 \\ & + \frac{E_{\max} \cdot z_1^{h_P}}{\text{EC}_{50}^{h_P} + z_1^{h_P}} - k_{\text{Dex}} \cdot z_1 \cdot z_2, \\\nonumber 
\end{aligned}
\end{equation*}
\begin{gather*}
\dot{z}_2 = -k_2 \cdot z_2 + k_3 \cdot z_3, \quad 
\dot{z}_3 = -k_3 \cdot z_3.
\end{gather*}
\[
\dot{z}_4 = k_{DP} \cdot z_4 - k_{IIR} \cdot z_4 \cdot z_1 - k_{DC} \cdot z_4 \cdot z_5^{h_C} \]
\[
\dot{z}_5 = k_1 \cdot z_1, \]

The parameters \( k_{IR} \), \( k_{PF} \), and \( k_O \) govern the innate immune response, modeling its stimulation by viral load, positive feedback, and natural decay, respectively. The term \( E_{\max} \) controls the immune activation saturation, with \( \text{EC}_{50} \) defining the half-maximal effective concentration. The suppression effect of dexamethasone on the immune response is captured by \( k_{\text{Dex}} \). The viral replication dynamics are determined by \( k_{DP} \), while \( k_{IIR} \) and \( k_{DC} \) regulate the effects of innate and adaptive immunity on viral clearance. The growth of adaptive immunity is parameterized by \( k_1 \). The pharmacokinetics of dexamethasone follow a two-compartment model, where \( k_2 \) and \( k_3 \) govern drug clearance and inter-compartmental transfer.

\textit{Expert ODE Adaptation (~\cite{leon2023modelling})}
Our expert ODE model removes \( z_5 \) (adaptive immunity) to focus on acute innate immune response dynamics. The revised system is:

\begin{equation*}
\begin{aligned}
\dot{z}_1 = & k_{IR} \cdot z_4 + k_{PF} \cdot z_4 \cdot z_1 - k_O \cdot z_1 \\ & + \frac{E_{\max} \cdot z_1^{h_P}}{\text{EC}_{50}^{h_P} + z_1^{h_P}} - k_{\text{Dex}} \cdot z_1 \cdot z_2, \\\nonumber
\end{aligned}
\end{equation*}
\[
\dot{z}_4 = k_{DP} \cdot z_4 - k_{IIR} \cdot z_4 \cdot z_1 - k_{DC} \cdot z_4 \]
\[
\dot{z}_2 = -k_2 \cdot z_2 + k_3 \cdot z_3 \qquad 
\dot{z}_3 = -k_3 \cdot z_3 \]

In this model, \( z_1 \) represents the innate immune response to infection, while \( z_4 \) tracks viral load dynamics. The variables \( z_2 \) and \( z_3 \) describe the pharmacokinetics of dexamethasone, where \( z_3 \) represents plasma concentration and \( z_2 \) describes lung tissue concentration. Removing \( z_5 \) simplifies the model to focus on the direct interaction between dexamethasone and innate immunity, emphasizing acute interventions.

\textit{Treatment Modeling.}
The administration of dexamethasone follows a pharmacokinetics-driven decay model. The plasma concentration \( z_3(t) \) is given by:

\begin{equation}
z_3(t) = \sum_{i} k_d \cdot d_i \cdot I(t > t_i) \cdot \exp(k_3 (t_i - t)).
\end{equation}

where \( k_d = 5 \) is a scaling factor ensuring that \( d_i \) remains within \([0,1]\), correctly weighting the administered dose. The variable \( d_i \) represents the dose given at time \( t_i \), while \( I(t > t_i) \) is an indicator function that activates the dose effect once administered. This formulation ensures that dexamethasone concentration follows a two-compartment decay model, where drug elimination is governed by \( k_3 \). The treatment effect propagates through \( k_{\text{Dex}} \), which modulates immune response suppression.

\section{Comparison between the expert ODE and Data Generation ODEs}
As previously mentioned, our method is designed to leverage imperfect mechanistic models to provide effective diffusion guidance. To illustrate this, we present a direct comparison between the ODEs used for data generation and those used during model inference in Table 3. It is important to emphasize again that we deliberately excluded access to the data generation ODE during inference to ensure that the models are indeed imperfect and reflect the realities of real-world scenarios.

As shown clearly in the table, the expert ODE lacks several components present in the data generation ODE, which significantly reduces its capacity to capture the underlying dynamics. Notably, in the Dexamethasone experiment, even though the expert ODE omits only a single variable $z_5$, this omission has a substantial impact on its ability to model the system’s behavior~\cite{leon2023modelling}.

\begin{table*}[]
    \centering
    \begin{tabular}{c|c|c}
    \toprule
       Dataset  & Data generation ODEs & Expert ODEs \\
    \midrule
        Covid-19 &   \makecell{$\dot{S} = -d_{S,E},$ \\ $\dot{E}=d_{S,E}-d_{E,I_P},$ \\ $\dot{I}_{A} = d_{E,IA} - d_{IA,R},$ \\ $\dot{I}_{P} = d_{E,IP} \textcolor{red}{- d_{IP,IS} - d_{IP,IM}}, $ \\
        \textcolor{red}{$\dot{I}_{M} = d_{IP,IM} - d_{IM,R},$} \\
        \textcolor{red}{$\dot{I}_{S} = d_{IP,IS} - d_{IS,HR} - d_{IS,HD}, $} \\
        \textcolor{red}{$\dot{H}_{R} = d_{IS,HR} - d_{HR,R},$}\\
        \textcolor{red}{$\dot{H}_{D} = d_{IS,HD} - d_{HD,D},$}\\
        $\dot{R} = d_{IA,R} \textcolor{red}{+ d_{IM,R} + d_{HR,R}}, $\\
        $\dot{D} = d_{HD,D},$
        } &
        \makecell{$\dot{S} = - \beta S / N, $\\ $\dot{E}  = \beta S / N - \alpha E ,$ \\
$\dot{I} = \alpha E  - \gamma I - \mu I ,$\\ $\dot{R} = \gamma I ,$\\ $\dot{M} = \mu I$}\\
        \midrule
        Dexamethasone & \makecell{$
            \begin{aligned}
                \dot{z}_1 = & k_{IR} \cdot z_4 + k_{PF} \cdot z_4 \cdot z_1 - k_O \cdot z_1 \\ & + \frac{E_{\max} \cdot z_1^{h_P}}{\text{EC}_{50}^{h_P} + z_1^{h_P}} - k_{\text{Dex}} \cdot z_1 \cdot z_2
            \end{aligned}
        $, \\
$\dot{z}_2 = -k_2 \cdot z_2 + k_3 \cdot z_3, $\\ 
$\dot{z}_3 = -k_3 \cdot z_3, $\\
$\dot{z}_4 = k_{DP} \cdot z_4 - k_{IIR} \cdot z_4 \cdot z_1 - k_{DC} \cdot z_4\textcolor{red}{ \cdot z_5^{h_C}},$\\
\textcolor{red}{$\dot{z}_5 = k_1 \cdot z_1$}} & \makecell{$
            \begin{aligned}
                \dot{z}_1 = & k_{IR} \cdot z_4 + k_{PF} \cdot z_4 \cdot z_1 - k_O \cdot z_1 \\ & + \frac{E_{\max} \cdot z_1^{h_P}}{\text{EC}_{50}^{h_P} + z_1^{h_P}} - k_{\text{Dex}} \cdot z_1 \cdot z_2
            \end{aligned}
        $, \\
$\dot{z}_2 = -k_2 \cdot z_2 + k_3 \cdot z_3, $\\ 
$\dot{z}_3 = -k_3 \cdot z_3, $\\
$\dot{z}_4 = k_{DP} \cdot z_4 - k_{IIR} \cdot z_4 \cdot z_1 - k_{DC} \cdot z_4$}\\
        \bottomrule
    \end{tabular}
    \caption{Comparison between the expert ODE and the data generation ODE, where the red components indicate the parts included in the data generation ODE but omitted from the expert ODE.}
    \label{tab:placeholder}
\end{table*}

\section{Denoising Diffusion Probabilistic Models (DDPM)}
\label{sec:DDPMs}
A standard diffusion model is defined by a forward process $q$ and a reverse process $p$. In the forward process, Gaussian noise is gradually added to the data, following a Markov chain:
\begin{equation*}
\begin{aligned}
    &q(y_t|y_{t-1}):=\mathcal{N}(y_t;\sqrt{1-\beta_t} y_{t-1},\beta_t \mathbf{I}),\\
    &q(y_{1:T}|y_0):=\prod_{t=1}^Tq(y_t|y_{t-1})
\end{aligned}
\end{equation*}
where $\{\beta_t\in(0,1)\}_{t=1}^T$ is the variance schedule. By leveraging the reparameterization trick, a sample $y_t$ at any given time $t$ can be expressed as:
\begin{equation*}
    q(y_t|y_0)=\mathcal{N}(y_t;\sqrt{ \Bar{\alpha}_t}y_0,(1-\Bar{\alpha}_t\mathbf{I}))
\end{equation*}
where $\alpha_t:=1-\beta_t$ and $\Bar{\alpha}_t:=\prod_{s=1}^t\alpha_s$. Finally $y_t$ is equivalent to a Gaussian noise when $T\rightarrow\infty$.

To reconstruct data that has been corrupted into Gaussian noise, we need to reverse the forward process and sample from $q(x_{t-1}|x_t)$. When the noise level $\beta_t$ is small enough, $q(x_{t-1}|x_t)$ can also be considered a Gaussian process. However, this process is difficult to model directly. Therefore, diffusion models $p_\theta$ have been proposed to approximate this conditional probability and reconstruct the original data:
\begin{equation*}
\begin{aligned}
    &p_\theta(\mathbf{y}_{0:T})=p(\mathbf{y_T})\prod^T_{t=1}p_\theta(y_{t-1}|y_t) \\
    &p_\theta(y_{t-1}|y_t)=\mathcal{N}(y_{t-1};\mu_\theta(y_t,t),\Sigma_\theta(y_t,t))
\end{aligned}
\end{equation*}

Since directly computing and optimizing the likelihood $p(y_0)$ is challenging due to intractable integration, we consider the following parameterization:
\begin{equation*}
    \Tilde{\mu}_t = \frac{1}{\sqrt{\alpha_t}}(y_t-\frac{1-\alpha_t}{\sqrt{1-\Bar{\alpha_t}}}\epsilon_t)
\end{equation*}
and derive the commonly used training loss function in practice through variational inference and simplifications:
\begin{equation}
    L_t = \mathbb{E}_{t\sim[1,T],y_0,\epsilon_t}[\|\epsilon_t-\epsilon_\theta(\sqrt{\Bar{\alpha}_t}y_0+\sqrt{1-\Bar{\alpha}_t}\epsilon_t,t)\|^2]
\end{equation}





\section{Factual guidance.}
To enhance the accuracy and reliability of counterfactual prediction by leveraging expert knowledge from counterfactual qualitative state estimation and the ground truth factual values, ODE-Diff employs classifier guidance for further calibration during conditional generation. Specifically, we introduce two loss terms: the counterfactual qualitative loss and the factual loss. The counterfactual qualitative loss $\text{Loss}_{\text{cf}}$ is directly defined by Equation (). The factual loss $\text{Loss}_{\text{f}}$ is derived from the consistency between the counterfactual and factual data observed prior to the change in treatment, which is formally written as:
\begin{equation}
    \text{Loss}_{\text{f}} = \sum_{t\in\Gamma^a}\|y_0^t-\hat{y}_0^t(y_\tau,\tau,\theta|y',x,a)\|_2^2
\end{equation}
where $\Gamma^a$ denotes the timeframe prior to the policy change, $y_0^t$ and $\hat{y}_0^t(y',x,a,\theta)$ represent the ground truth factual time-series value at time $t$ and generated counterfactual time-series value at time $t$, respectively.

At each diffusion step $\tau$, we directly update the time-series prediction $\hat{y}_0(y_\tau,\tau,\theta|y',x,a)$ according to Equation (6), and compute $\text{Loss}_{\text{cf}}$ and $\text{Loss}_{\text{f}}$ based on the updated prediction, enabling efficient guidance of the diffusion model:
\begin{equation}
\begin{aligned}
    \Tilde{y}_0(y_\tau,\tau,\theta|y',x,a) & =  \hat{y}_0(y_\tau,\tau,\theta|y',x,a) \\ & + \eta\nabla_{y_\tau}\text{Loss}_{\text{cf}} + \nu\nabla_{y_\tau}\text{Loss}_{\text{f}}
\end{aligned}
\end{equation}
where $\eta$ and $\nu$ are guidance strength parameters for counterfactual qualitative guidance and factual guidance.

\section{Extended Experimental Setup}

\subsection{Metrics}

The k-Wasserstein distance measures the distance between two distributions $\nu$ and $\mu$, formally: 
\[W^k(\nu,\mu)=(\int_0^1|\mathbb{F}_1^{-1}(l)-\mathbb{F}_2^{-1}(l)|^k)^\frac{1}{k}\]
where $\mathbb{F}_1^{-1}(l)$ and $\mathbb{F}_2^{-1}(l)$ represent the inverse cumulative distribution functions (CDFs) of $\nu$ and $\mu$ for quantile $l$. In our experiments, we report the $W^1$ distance and lower values are preferred. 

Since generating potential outcome distributions naturally provides advantages in uncertainty quantification, we evaluate \ourmethod's ability to learn causal relationships and quantify uncertainty by assessing the predictive intervals of its generated counterfactual distributions~\cite{hess2023bayesian}. Specifically, we select commonly used confidence levels of 75\%, 90\% and 95\% and measure the frequency at which the generated prediction intervals cover the test data. This serves as an indicator of the accuracy of uncertainty quantification. Ideally, the coverage frequency of the prediction intervals should closely match the specified confidence levels, indicating the model's reliability in capturing uncertainty.

Since our method directly outputs the potential outcome distribution, we can use it as an intermediate variable to compute the Conditional Average Treatment Effect (CATE), highlighting the versatility of our approach. Specifically, we first compute the mean of the generated potential outcomes $\mathbb{E}[Y|X,A=a]$ for all $a$ and then obtain the CATE value directly using its definition $\mathbf{CATE}(x)=\mathbb{E}[Y|X,1]-\mathbb{E}[Y|X,0]$. We compare the generated CATE with the ground truth, where a smaller RMSE indicates better performance.

\section{Extended Datasets}

\subsubsection{Semi-synthetic COVID-19 Data.} 
To investigate the impact of mask mandates across 121 metropolitan areas in the United States, we constructed a semi-synthetic dataset based on pandemic dynamics simulated using the SEIR-HD model \cite{kain2021chopping}. Population estimates were derived from the U.S. Census Bureau \cite{uscensus2020}, and the dataset spans 52 weeks from 2020 to 2021, with variables aggregated at a weekly resolution. The dataset includes two city-level covariates: new hospitalizations and infectious symptomatic cases. The treatment variable is the mask mandate policy (with values of 0 indicating no mandate and 1 indicating a mandate), and the outcome variable is the city-level death cases per 1000 people. Cities were classified into strict and relaxed policy groups based on the comprehensiveness and duration of their mandates \cite{nguyen2021impact}. To capture real-world variability, the simulation incorporates heterogeneity in hospitalization rates and transmission dynamics.  
We employ the SEIR-HD model as the data generation ODE to simulate the ground truth data. Unlike the SEIRM model, which captures basic disease transmission dynamics, SEIR-HD is a more complex epidemiological model that explicitly models hospitalization and disease severity for more realistic epidemic progression. \cite{kain2021chopping}. 

To emphasize our model's ability to perform causal modeling in environments with sparse dynamic knowledge and limited data, we rely solely on the population size of each city as the only real-world data input. To ensure realistic dynamics in the absence of rich observational data, we simulate different age structures and hospitalization resources across cities by adjusting parameters in the ground truth ODE, specifically modifying $\alpha$ and $\delta$ (details provided in \textit{Parameters Assignment}). 
Initial conditions are derived from census data \cite{uscensus2020} and epidemiological estimates of infectious/exposed ratios \cite{gandhi2021first}. Most strict-policy cities adopted mandates around week 15 \cite{chernozhukov2021causal}. The dataset includes 121 factual training samples and 121 counterfactual test samples, supporting robust evaluation of policy effects. Following content details classification criteria, enforcement sources, and simulation parameters.

\textit{Initial conditions} at $t=0$ were set based on fixed ratios of exposed, infectious, and hospitalized individuals relative to the total population, following early epidemiological studies \cite{gandhi2021first}. Specifically, we assigned:

\begin{table*}[h]
    \centering
    \renewcommand{\arraystretch}{1.2}
    \caption{Initial distribution of COVID-19 compartments in the SEIR-HD model.}
    \begin{tabular}{l l l l}
        \hline
        \textbf{Category} & \textbf{Value} & \textbf{Category} & \textbf{Value} \\ 
        \hline
        Exposed & 0.15\% of total population & Infectious Asymptomatic & 0.1\% \\ 
        Infectious Pre-symptomatic & 0.07\% & Infectious Mild & 0.05\% \\ 
        Infectious Severe & 0.02\% & Hospitalized Recovered & 0.001\% \\ 
        Hospitalized Deceased & 0.0005\% & Recovered & 0.00005\% \\ 
        Deceased & 0.00001\% & & \\ 
        \hline
    \end{tabular}
\end{table*}

The susceptible population was then computed as the remainder after assigning these proportions.

\textit{Mask mandate policies} were categorized into \textit{strict} (1) and \textit{relaxed} (0) groups based on real-world enforcement data \cite{nguyen2021impact}. Cities implementing strict mandates generally did so around week 15, while late adopters enforced mandates around week 40 \cite{chernozhukov2021causal}. This classification enables structured counterfactual analysis of mask mandate impacts on public health.

\textit{Parameters Assignment.}  
Following \cite{kain2021chopping}, we utilize the SEIR-HD model, which extends the SEIRM model by incorporating additional compartments to capture hospitalization and disease severity. We adopt the parameter values from \cite{kain2021chopping} as the default setting. To reflect city-specific heterogeneity, we assign different values for the transition rates $\alpha$ and $\delta$ based on \cite{lyu2020community}, which accounts for differences in age distributions and healthcare resources across cities. For simplicity, given our focus on scarce data settings, we assign:
\begin{align*}
\text{Strict policy:} \quad & \delta = 0.15, \quad \alpha = 0.3, \\
\text{Relaxed policy:} \quad & \delta = 0.1, \quad \alpha = 0.5.
\end{align*}

\textit{Simulating the Effects of Interventions.}  
To model the impact of mask mandates, we simulate the effect of interventions by introducing a decay mechanism for the contact rate $\beta$. Since public policies such as mask mandates restrict population movement, they effectively reduce the average contact rate, thereby lowering transmission. We define the baseline transmission rate as $\beta = 0.5$ and apply an exponential decay schedule when mandates are enforced:
\begin{align*}
\beta_{\text{values}} = \texttt{get\_beta\_schedule}(&\text{initial\_beta} = 0.5,\\ & \lambda = 0.005).
\end{align*}

This schedule gradually reduces $\beta$ over time, simulating the progressive effects of policy interventions on disease transmission.

\subsubsection{Fully-synthetic Dexamethasone Data.}  
In this simulation, we use mechanistic models to predict the results of a single dexamethasone treatment. The simulation involves 50 patients, and we consider a time horizon of $T = 14$ days, which corresponds to the median length of hospital stay for COVID-19 patients. We use a pharmacological model adapted from \cite{dai2021prototype} that describes five expert variables ($E = 5$) under dexamethasone treatment for COVID-19 patients. This is a complex version of an ODE and includes a supplementary variable for the adaptive immune response compared to the PKPD pharmacological model described in Section 3. The detailed model structure and the expert variables are specified in section \textit{The pharmacological model for dexamethasone}.

The treatment variable represents the dosage $d_i \in [0,1]$, with a scaling parameter $k_d = 5$ to represent the average dose level \cite{qian2021integrating}. To simulate the real-world scarcity of observed data for each patient, we simplify the treatment timing to $t=3$, reflecting the empirical observation that patients typically receive a dose around the third day \cite{Horby2021Dexamethasone}. In the training set, we randomly assign half of the patients to receive a one-time treatment, while the remaining patients receive no treatment dose. The testing set consists of 50 counterfactual cases, allowing for the evaluation of treatment effects. The ground truth ODE's $z_1$, representing the innate immune response, serves as the output measurement. The covariate $x$ is a physiological variable that integrates the five expert variables from the ground truth ODE through a linear layer (Appendix B), typically denoting physiological markers such as reactive protein levels.

For each patient $i$, each component of the initial condition $z_i(0)$ is independently drawn from an exponential distribution with rate $\lambda = 100$ (more details in Appendix B). Measurement noises are independently drawn from $\epsilon_{it} \sim \mathcal{N}(0, \sigma^2)$, with $\sigma = 0.01$. To emphasize our model's ability to perform causal modeling in environments characterized by sparse dynamic knowledge and limited data, we utilize fixed dosing times and simplified treatment protocols to reflect real-world data scarcity. Simultaneously, we incorporate empirical values to ensure the realism of the physiological responses, enabling robust counterfactual analysis of dexamethasone treatment impacts on COVID-19 patients.

\textit{Dose and Timing Justification.} The dosage $d_i = 5$ is selected to represent the average dose typically administered to patients, as the standard range lies between $[0,10]$ \cite{qian2021integrating}. We simplify this by taking the midpoint. The treatment time $t=3$ is chosen based on empirical observations that dexamethasone reduces mortality in COVID-19 patients requiring oxygen or mechanical ventilation, typically initiated around the third day of hospitalization when respiratory distress manifests \cite{Horby2021Dexamethasone}.

\textit{Covariate and Measurement Generation.} The covariate $x_i$ is generated as:
\begin{align*}
    x_i = \mathbf{W}_3 z_i + \mathbf{W}_4 a_i,
\end{align*}
with coefficient matrices $\mathbf{W}_3 \in \mathbb{R}^{X \times (M+E)}$ and $\mathbf{W}_4 \in \mathbb{R}^{X \times 1}$. Each element in these matrices is drawn independently from $\mathcal{N}(0, 1)$ and multiplied by a Bernoulli variable with $p=0.5$, ensuring that approximately half of the elements in $\mathbf{W}_3$ and $\mathbf{W}_4$ are zero. This reflects the idea that each physiological variable is only related to some latent variables. Unlike other models that might generate multiple covariates, we integrate all expert variables and dosage information into a single $x$ to simplify the representation, ensuring that it captures the holistic physiological response while maintaining realism.

The measurement $y_i(t)$, which is the predicted variable, is directly derived from $z_1(t)$ representing the innate immune response. To enhance realism, we add measurement noise $\epsilon_{it} \sim \mathcal{N}(0, 0.01)$, leading to the following measurement equation:
\begin{align*}
    y_i(t) = z_1(t) + \epsilon_{it}.
\end{align*}

We first simulate all the daily measurements at $t = 1, 2, \ldots, T$, and then randomly remove measurements with probability 0.5; this represents the fact that measurements are made irregularly.

\textit{Expert Variables Initialization.} The initial conditions for expert variables in the ground truth ODE are detailed as follows. For each patient $i$, components $z_2(0)$ and $z_3(0)$ are drawn from an exponential distribution with rate $\lambda = 100$, representing initial states at ICU admission where dexamethasone levels are minimal. Conversely, $z_1(0)$, $z_4(0)$, and $z_5(0)$, which represent immune responses and viral loads, are drawn from an exponential distribution with rate $\lambda = 0.1$ to allow for greater heterogeneity among patients. These distributions not only model the variability in patient responses but also ensure the positivity of the expert variables, which aligns with their biological interpretation.

\section{Selection of Guidance Strength}
As we previously mentioned, a key challenge in classifier-guided diffusion models for counterfactual estimation lies in determining the appropriate guidance strength $\eta$. To address this, we fully leverage the domain knowledge provided by the expert ODE and design a heuristic two-step procedure: first, we align the factual prediction of the diffusion model with the factual simulation from the expert ODE and apply the same alignment to the diffusion model’s counterfactual prediction. Then, we compute the correlation between the difference of the diffusion model’s counterfactual and factual predictions under different values of $\eta$, and the corresponding difference from the expert ODE. We select the value of $\eta$ that yields the highest correlation. This method effectively utilizes domain knowledge while minimizing errors introduced by the rigidity of mechanistic models. Here, we first report the selected guidance strength $\eta$ values on the two datasets, followed by performance curves and correlation curves on the two synthetic datasets to visually demonstrate the validity of this approach.
\begin{table}[]
    \centering
    \begin{tabular}{c|cc}
    \toprule
        Value & Value $\eta$ & Directional $\eta$\\
        \midrule
       Covid-19 Data  & 2000 & 2000 \\
       Dexamethasone Data  & 1000 & 200 \\ 
       \bottomrule
    \end{tabular}
    \caption{The selected value for guidance strength $\eta$ for both value guidance and directional guidance on Covid-19 and Dexamethasone datasets}
    \label{tab:placeholder}
\end{table}
\begin{figure*}
    \centering
    \includegraphics[width=1.0\linewidth]{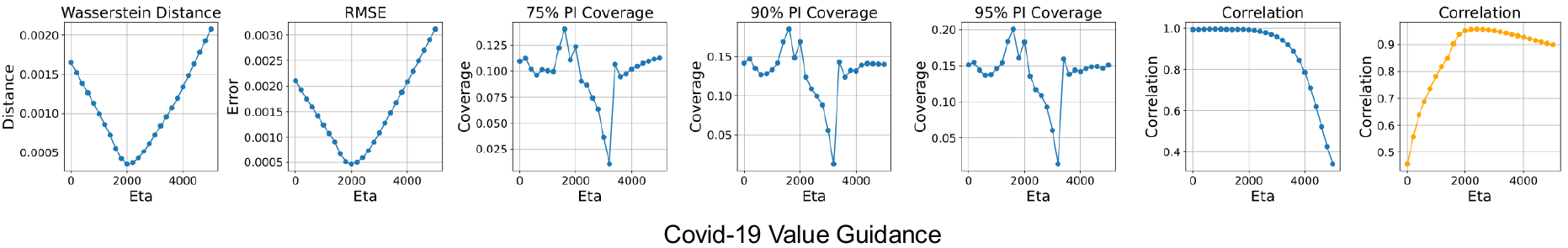}
    \includegraphics[width=1.0\linewidth]{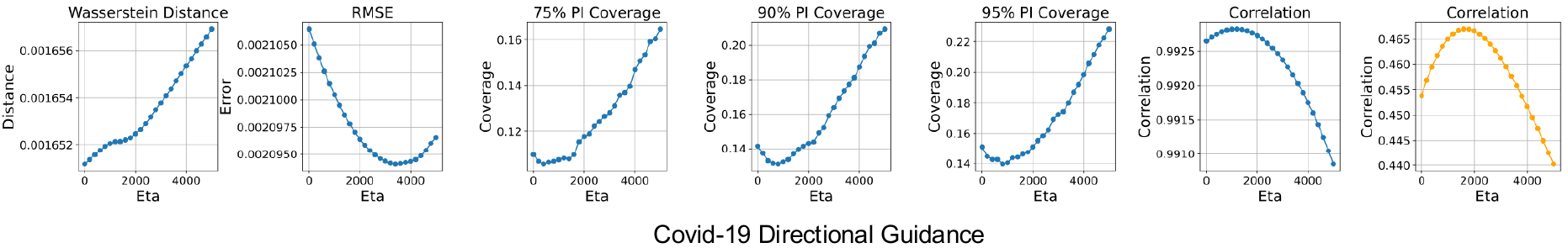}
    \includegraphics[width=1.0\linewidth]{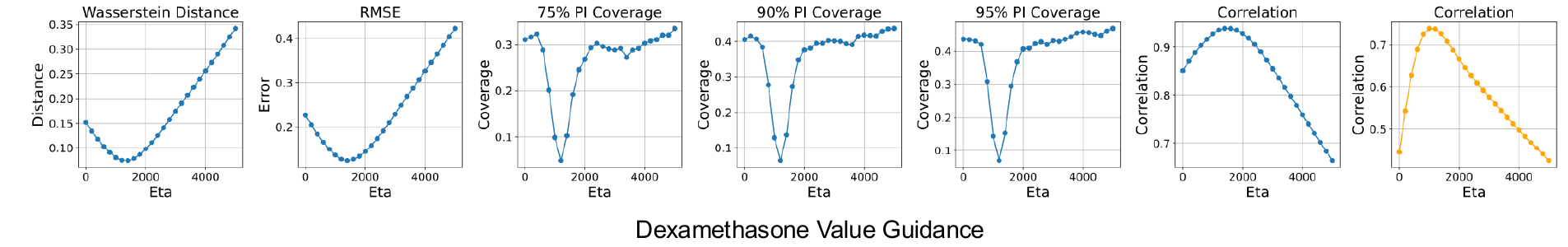}
    \includegraphics[width=1.0\linewidth]{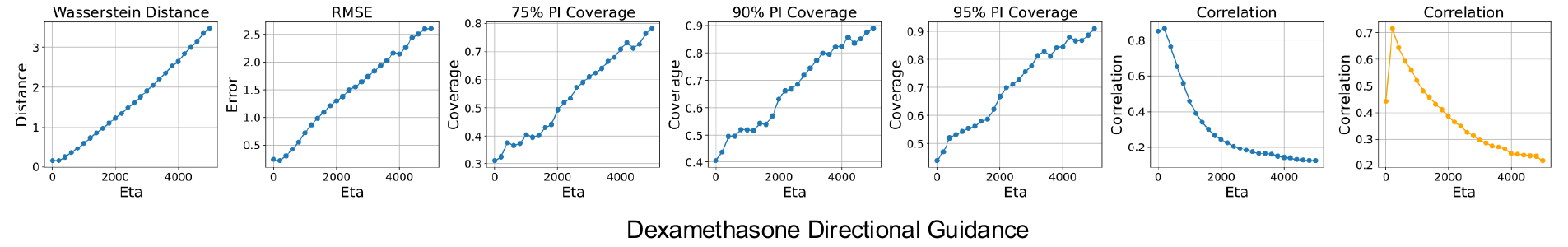}
    \caption{The metric performance and correlation values corresponding to different guidance strengths. The blue curve represents the performance metric, while the orange curve denotes the selection criterion (correlation).}
    \label{fig:placeholder}
\end{figure*}

As shown in Figure 3, the $\eta$ selected by our heuristic algorithm corresponds to a value that yields reasonable performance across evaluation metrics.

\section{Supplemental Details for the Real-World Case Study}

On the U.S. state level, we use a Dynamic Time Warping (DTW) algorithm to find temporally similar regions for each state during the training period (defined as the first 25 epiweeks). The aim is to identify peer states with comparable early dynamics but diverging policy responses in the latter half of the timeline. These matched states are used to construct a synthetic control by averaging their outcomes, providing a more realistic counterfactual trajectory for each target state.

\vspace{1em}

\begin{figure*}[h]
    \centering
    \includegraphics[width=0.9\textwidth]{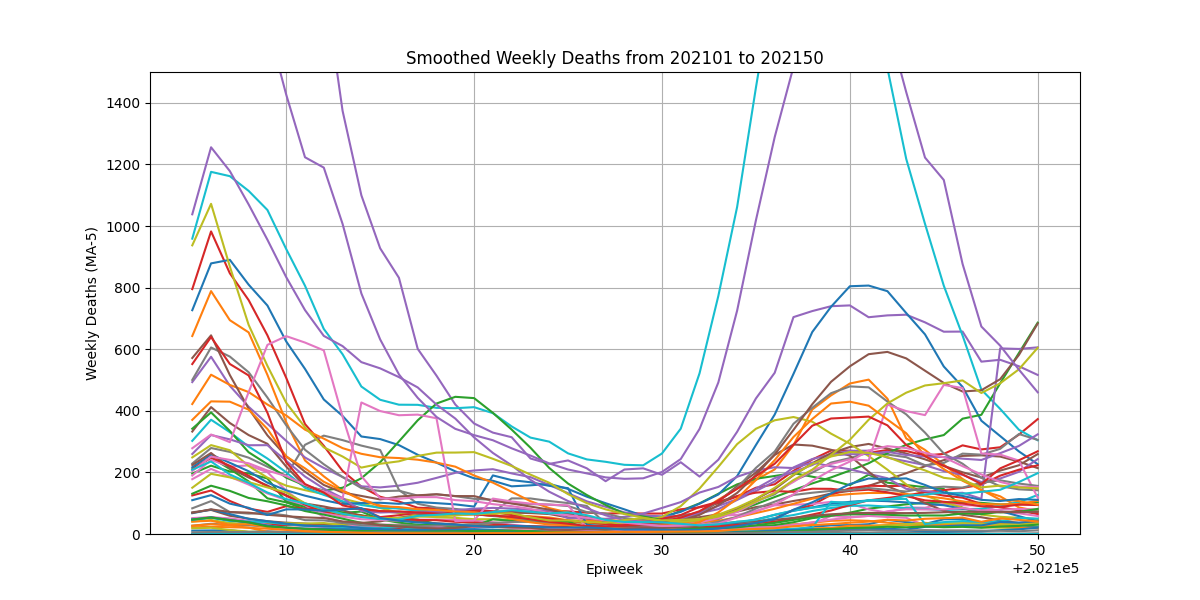}
    \caption{Smoothed weekly deaths (MA-5) from \texttt{start\_week} to \texttt{end\_week} across all regions.}
    \label{fig:weekly_deaths}
\end{figure*}

\begin{figure*}[h]
    \centering
    \includegraphics[width=0.9\textwidth]{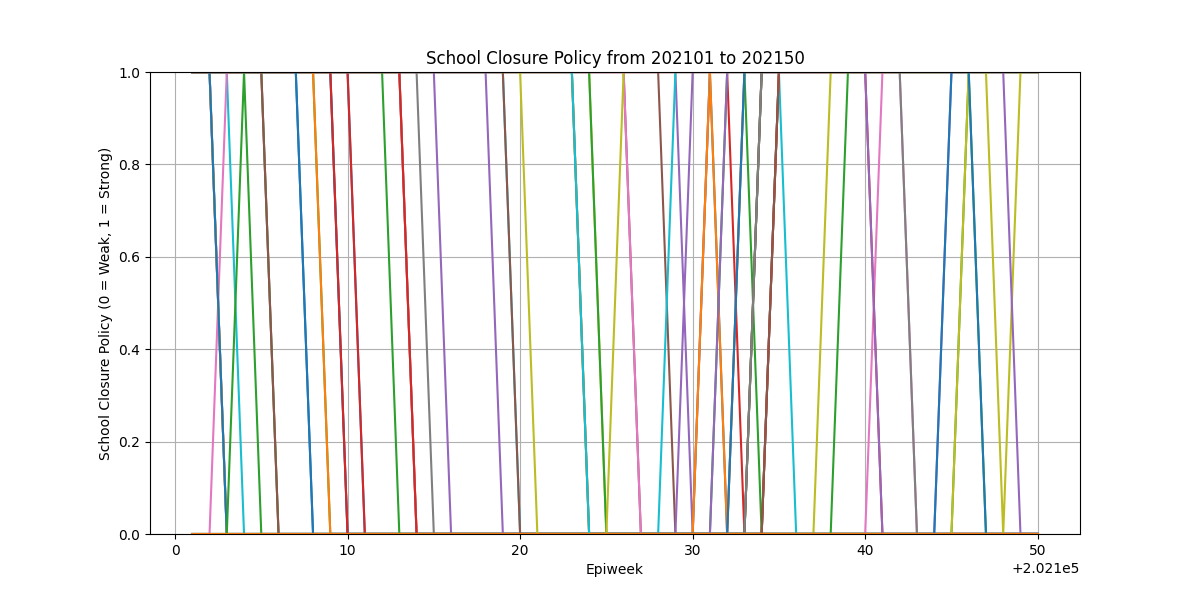}
    \caption{School closure policies from \texttt{start\_week} to \texttt{end\_week} across all regions. Binary levels (0 = weak or absent, 1 = strong) reflect state-level intervention strictness.}
    \label{fig:school_closures}
\end{figure*}

Figure~\ref{fig:weekly_deaths} shows the smoothed weekly COVID-19 deaths (5-epiweek moving average) for all regions considered in our analysis, while Figure~\ref{fig:school_closures} displays the corresponding school closure policy levels across the same time range. These figures allow us to visually inspect the policy heterogeneity and outcome variability across states.

The variability shown in these plots highlights the feasibility of constructing counterfactual estimates using matched regions. For instance, even if two states exhibited similar early trajectories in terms of weekly deaths, differences in intervention policies (e.g., school closures) in the later period enable us to learn causal effects by contrasting their outcomes. This provides an empirical basis for evaluating both our model and its counterfactual outputs.

Our model evaluates performance using both quantitative metrics—such as the Wasserstein Distance (WD) between predicted and actual outcomes—and qualitative assessments through visualizations. The reduction in WD between factual and counterfactual distributions is indicative of our model’s ability to distinguish between policy regimes.

\subsection{Comparison to Prior Policy Impact Work}

\begin{figure*}[t]
    \centering
    \includegraphics[width=0.75\linewidth]{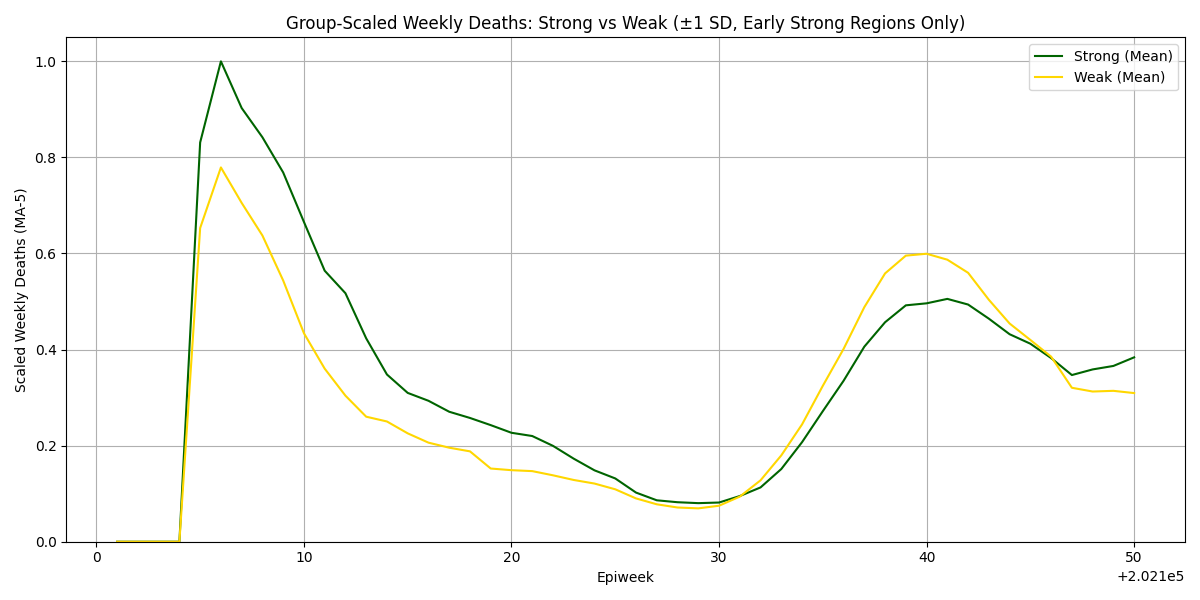}
    \caption{
        Comparison of states starting with strong school closure policies. States that later weakened policies in the test period show higher mortality peaks, while those maintaining strong policies consistently observed lower peaks. This highlights the value of sustained interventions.
    }
    \label{fig:strong_vs_weak}
\end{figure*}

Our findings are consistent with prior large-scale evaluations of non-pharmaceutical interventions (NPIs), including the Nature paper by Arık et al.~\cite{arik2021prospective}. In their work, they assess the impact of AI-augmented models on epidemic forecasting in the U.S. and Japan, noting the strong influence of early and consistent policies on downstream mortality and case dynamics.

Our approach, while distinct in methodology, leads to similar conclusions: targeted and early interventions such as school closures have observable effects on COVID-19 trends, especially when regional heterogeneity is leveraged. As shown in Figure~\ref{fig:strong_vs_weak}, when our policy recommendations are mapped across states, we observe patterns similar to those in~\cite{arik2021prospective}, underscoring the effectiveness of region-specific adaptive policymaking.

These results suggest that using our counterfactual framework for policy design could aid in identifying which interventions would be most effective for a given region—an insight that can be operationalized in future public health planning.

\subsection{Case-Study Experimental Procedure}
In this case study, we analyze weekly COVID-19 death counts and school-closure policies for all 52 U.S.\ regions.  Since no true counterfactual exists, our procedure is:

\begin{enumerate}
  \item \textbf{Similarity \& Proxy Counterfactual.}  
    Compute Dynamic Time Warping (DTW) over the first 25 weeks to identify, for each region, its five closest neighbors by trajectory.  These neighbors serve as a stand-in counterfactual repository.
  \item \textbf{Train/Test Split.}  
    Randomly select 10 regions as a \emph{test set} and use the remaining 42 for training.  Retrain the model for each fold to ensure no data leakage.
  \item \textbf{Policy Group Labeling.}  
    Label each test region and its five neighbors as “strong” or “weak” school-closure based on the dominant policy regime during the post-flip period.
  \item \textbf{Evaluation via Wasserstein Distance (WD).}  
    \begin{itemize}
      \item \emph{Proxy} $\mathrm{WD}_{\mathrm{neigh}}$: the Wasserstein distance between the \emph{average} death-rate curves of the strong vs.\ weak neighbor groups.
      \item \emph{Predicted} $\mathrm{WD}_{\mathrm{model}}$: force the test region’s policy to strong vs.\ weak after its first flip, predict both trajectories, and compute the WD between them.
      \item \emph{Accuracy Measure}: compare $\mathrm{WD}_{\mathrm{neigh}}$ vs.\ $\mathrm{WD}_{\mathrm{model}}$ for each region; smaller discrepancies indicate more faithful policy‐impact estimation.
    \end{itemize}
\end{enumerate}

This experimental setup closely mirrors real‐world public health decision‐support: officials observe only the factual time series and must infer counterfactuals from comparable jurisdictions.  Demonstrating small differences between proxy and predicted WDs validates our model’s ability to accurately forecast the effects of non‐pharmaceutical interventions, informing timely resource allocation and policy evaluation during an ongoing epidemic.  

\begin{table*}[ht]
\centering
\caption{Per‐region Wasserstein Distance comparison: nearest neighbor counterfactual ground‐truth proxy vs.\ model‐predicted $\Delta$ for GANITE and for TECDE.}
\begin{tabular}{lccccccc}
\toprule
\textbf{Region} & \textbf{Class.} & \textbf{ProxyGT} & \textbf{GANITE $\Delta$} & \textbf{TECDE $\Delta$}  &\textbf{MS-Diffusion $\Delta$}  & \textbf{DiffPO $\Delta$} & \textbf{\ourmethod $\Delta$} \\
\midrule
FL & weak   & 0.1434 & 0.3825 & 0.0338 & 0.0056 & 0.0241 & 0.0336 \\
MN & weak   & 0.1908 & 0.3749 & 0.0155 & 0.0223 & 0.0175 & 0.0375 \\
MS & weak   & 0.2015 & 0.3763 & 0.0311 & 0.0135 & 0.0183 & 0.0314 \\
AZ & weak   & 0.0356 & 0.3345 & 0.0188 & 0.0127 & 0.0312 & 0.0325 \\
ME & strong & 0.1762 & 0.3955 & 0.0213 & 0.0077 & 0.0103 & 0.0377 \\
PA & weak   & 0.1171 & 0.3465 & 0.0162 & 0.0158 & 0.0144 & 0.0328 \\
KS & weak   & 0.1423 & 0.3576 & 0.0187 & 0.0112 & 0.0115 & 0.0396 \\
MA & strong & 0.1414 & 0.3244 & 0.0197 & 0.0116 & 0.0146 & 0.0336 \\
HI & strong & 0.0631 & 0.3959 & 0.0328 & 0.0095 & 0.0116 & 0.0306 \\
VA & strong & 0.1024 & 0.3387 & 0.0308 & 0.0175 & 0.0120 & 0.0365 \\
MO & weak   & 0.2389 & 0.3595 & 0.0175 & 0.0117 & 0.0104 & 0.0410 \\
NV & weak   & 0.1044 & 0.3881 & 0.0191 & 0.0152 & 0.0155 & 0.0347 \\
NE & weak   & 0.2148 & 0.3555 & 0.0127 & 0.0081 & 0.0192 & 0.0372 \\
NJ & strong & 0.2497 & 0.3234 & 0.0324 & 0.0093 & 0.0177 & 0.0369 \\
TN & weak   & 0.2051 & 0.3644 & 0.0238 & 0.0097 & 0.0255 & 0.0348 \\
CA & strong & 0.2122 & 0.3180 & 0.0310 & 0.0068 & 0.0153 & 0.0294 \\
ND & weak   & 0.1569 & 0.3780 & 0.0173 & 0.0322 & 0.0203 & 0.0373 \\
GA & weak   & 0.2482 & 0.3828 & 0.0050 & 0.0177 & 0.0127 & 0.0382 \\
MT & weak   & 0.1127 & 0.3911 & 0.0159 & 0.0193 & 0.0397 & 0.0338 \\
OK & weak   & 0.1060 & 0.3649 & 0.0150 & 0.0312 & 0.0110 & 0.0378 \\
AK & weak   & 0.1173 & 0.3740 & 0.0128 & 0.0085 & 0.0153 & 0.0368 \\
TX & strong & 0.1242 & 0.3767 & 0.0182 & 0.0131 & 0.0325 & 0.0357 \\
LA & strong & 0.0730 & 0.3741 & 0.0230 & 0.0152 & 0.0179 & 0.0367 \\
SD & weak   & 0.0680 & 0.3417 & 0.0215 & 0.0114 & 0.0191 & 0.0354 \\
WV & weak   & 0.0924 & 0.3790 & 0.0107 & 0.0067 & 0.0163 & 0.0392 \\
OH & strong & 0.1388 & 0.3860 & 0.0124 & 0.0076 & 0.0229 & 0.0350 \\
WA & strong & 0.1836 & 0.3962 & 0.0110 & 0.0141 & 0.0188 & 0.0362 \\
NH & weak   & 0.0764 & 0.3468 & 0.0221 & 0.0146 & 0.0072 & 0.0350 \\
MD & strong & 0.1367 & 0.3380 & 0.0238 & 0.0121 & 0.0114 & 0.0399 \\
MI & strong & 0.1357 & 0.3395 & 0.0125 & 0.0089 & 0.0197 & 0.0347 \\
IA & weak   & 0.1274 & 0.3277 & 0.0177 & 0.0273 & 0.0285 & 0.0326 \\
WY & weak   & 0.0537 & 0.3812 & 0.0059 & 0.0112 & 0.0265 & 0.0438 \\
NM & strong & 0.1117 & 0.3380 & 0.0228 & 0.0093 & 0.0074 & 0.0397 \\
UT & strong & 0.2759 & 0.4037 & 0.0126 & 0.0115 & 0.0167 & 0.0377 \\
ID & weak   & 0.0485 & 0.3699 & 0.0073 & 0.0112 & 0.0119 & 0.0376 \\
SC & weak   & 0.3070 & 0.3597 & 0.0375 & 0.0287 & 0.0190 & 0.0338 \\
WI & weak   & 0.0707 & 0.3573 & 0.0103 & 0.0348 & 0.0164 & 0.0332 \\
AR & weak   & 0.3653 & 0.3827 & 0.0489 & 0.0173 & 0.0115 & 0.0347 \\
NC & weak   & 0.0840 & 0.3567 & 0.0427 & 0.0196 & 0.0114 & 0.0338 \\
OR & strong & 0.1901 & 0.4019 & 0.0439 & 0.0090 & 0.0132 & 0.0262 \\
CO & strong & 0.1060 & 0.3917 & 0.0420 & 0.0092 & 0.0130 & 0.0369 \\
CT & strong & 0.2273 & 0.3114 & 0.0321 & 0.0121 & 0.0211 & 0.0356 \\
DE & strong & 0.1596 & 0.3966 & 0.0291 & 0.0247 & 0.0169 & 0.0384 \\
IL & strong & 0.0674 & 0.3561 & 0.0159 & 0.0102 & 0.0128 & 0.0358 \\
IN & strong & 0.0576 & 0.3797 & 0.0170 & 0.0127 & 0.0086 & 0.0312 \\
KY & weak   & 0.2452 & 0.3930 & 0.0487 & 0.0157 & 0.0134 & 0.0324 \\
VT & strong & 0.1486 & 0.3903 & 0.0414 & 0.0101 & 0.0118 & 0.0347 \\
AL & weak   & 0.0968 & 0.3455 & 0.0250 & 0.0193 & 0.0174 & 0.0342 \\
DC & strong & 0.3306 & 0.2970 & 0.0493 & 0.0139 & 0.0263 & 0.0354 \\
RI & strong & 0.3336 & 0.3077 & 0.0323 & 0.0099 & 0.0230 & 0.0320 \\
\bottomrule
\end{tabular}
\end{table*}

%% file: aaai-main.bbl
\begin{thebibliography}{87}
\providecommand{\natexlab}[1]{#1}

\bibitem[{Alaa and Van Der~Schaar(2017)}]{alaa2017bayesian}
Alaa, A.~M.; and Van Der~Schaar, M. 2017.
\newblock Bayesian inference of individualized treatment effects using multi-task gaussian processes.
\newblock \emph{Advances in neural information processing systems}, 30.

\bibitem[{Arik et~al.(2021)Arik, Shor, Sinha, and et~al.}]{arik2021ai}
Arik, S.~{\"O}.; Shor, J.; Sinha, R.; and et~al. 2021.
\newblock A prospective evaluation of AI-augmented epidemiology to forecast COVID-19 in the USA and Japan.
\newblock \emph{npj Digital Medicine}, 4: 146.

\bibitem[{Ar{\i}k et~al.(2021)Ar{\i}k, Shor, Sinha, Yoon, Ledsam, Le, Dusenberry, Yoder, Popendorf, Epshteyn et~al.}]{arik2021prospective}
Ar{\i}k, S.~{\"O}.; Shor, J.; Sinha, R.; Yoon, J.; Ledsam, J.~R.; Le, L.~T.; Dusenberry, M.~W.; Yoder, N.~C.; Popendorf, K.; Epshteyn, A.; et~al. 2021.
\newblock A prospective evaluation of AI-augmented epidemiology to forecast COVID-19 in the USA and Japan.
\newblock \emph{NPJ digital medicine}, 4(1): 146.

\bibitem[{Augustin et~al.(2022)Augustin, Boreiko, Croce, and Hein}]{augustin2022diffusion}
Augustin, M.; Boreiko, V.; Croce, F.; and Hein, M. 2022.
\newblock Diffusion visual counterfactual explanations.
\newblock \emph{Advances in Neural Information Processing Systems}, 35: 364--377.

\bibitem[{Berrevoets et~al.(2021)Berrevoets, Curth, Bica, McKinney, and van~der Schaar}]{berrevoets2021disentangled}
Berrevoets, J.; Curth, A.; Bica, I.; McKinney, E.; and van~der Schaar, M. 2021.
\newblock Disentangled counterfactual recurrent networks for treatment effect inference over time.
\newblock \emph{arXiv preprint arXiv:2112.03811}.

\bibitem[{Bica et~al.(2020)Bica, Alaa, Jordon, and van~der Schaar}]{bica2020estimating}
Bica, I.; Alaa, A.~M.; Jordon, J.; and van~der Schaar, M. 2020.
\newblock Estimating counterfactual treatment outcomes over time through adversarially balanced representations.
\newblock In \emph{International Conference on Learning Representations}.

\bibitem[{Borchering et~al.(2023)Borchering, Healy, Cadwell, Johansson, Slayton, Wallace, and Biggerstaff}]{borchering2023public}
Borchering, R.~K.; Healy, J.~M.; Cadwell, B.~L.; Johansson, M.~A.; Slayton, R.~B.; Wallace, M.; and Biggerstaff, M. 2023.
\newblock Public health impact of the US Scenario Modeling Hub.
\newblock \emph{Epidemics}, 44: 100705.

\bibitem[{Chao, Bl{\"o}baum, and Kasiviswanathan(2023)}]{chao2023interventional}
Chao, P.; Bl{\"o}baum, P.; and Kasiviswanathan, S.~P. 2023.
\newblock Interventional and counterfactual inference with diffusion models.
\newblock \emph{arXiv preprint arXiv:2302.00860}, 4: 16.

\bibitem[{Chao et~al.(2024)Chao, Blöbaum, Patel, and Kasiviswanathan}]{chao2024modelingcausalmechanismsdiffusion}
Chao, P.; Blöbaum, P.; Patel, S.; and Kasiviswanathan, S.~P. 2024.
\newblock Modeling Causal Mechanisms with Diffusion Models for Interventional and Counterfactual Queries.
\newblock arXiv:2302.00860.

\bibitem[{Chen et~al.(2024)Chen, Li, Li, Wang, Wang, Liu, and Wang}]{chen2024quantifying}
Chen, K.; Li, G.; Li, H.; Wang, Y.; Wang, W.; Liu, Q.; and Wang, H. 2024.
\newblock Quantifying uncertainty: Air quality forecasting based on dynamic spatial-temporal denoising diffusion probabilistic model.
\newblock \emph{Environmental Research}, 249: 118438.

\bibitem[{Chen et~al.(2018)Chen, Rubanova, Bettencourt, and Duvenaud}]{chen2018neural}
Chen, R.~T.; Rubanova, Y.; Bettencourt, J.; and Duvenaud, D.~K. 2018.
\newblock Neural ordinary differential equations.
\newblock \emph{Advances in neural information processing systems}, 31.

\bibitem[{Chernozhukov, Fern{\'a}ndez-Val, and Melly(2013)}]{chernozhukov2013inference}
Chernozhukov, V.; Fern{\'a}ndez-Val, I.; and Melly, B. 2013.
\newblock Inference on counterfactual distributions.
\newblock \emph{Econometrica}, 81(6): 2205--2268.

\bibitem[{Chernozhukov, Kasahara, and Schrimpf(2021)}]{chernozhukov2021causal}
Chernozhukov, V.; Kasahara, H.; and Schrimpf, P. 2021.
\newblock Causal impact of masks, policies, behavior on early covid-19 pandemic in the U.S.
\newblock \emph{Journal of Econometrics}, 220(1): 23--62.
\newblock T=15.

\bibitem[{Curth and Van~der Schaar(2021)}]{curth2021nonparametric}
Curth, A.; and Van~der Schaar, M. 2021.
\newblock Nonparametric estimation of heterogeneous treatment effects: From theory to learning algorithms.
\newblock In \emph{International Conference on Artificial Intelligence and Statistics}, 1810--1818. PMLR.

\bibitem[{Dai et~al.(2021)Dai, Rao, Sher, Tania, Musante, and Allen}]{dai2021prototype}
Dai, W.; Rao, R.; Sher, A.; Tania, N.; Musante, C.~J.; and Allen, R. 2021.
\newblock A prototype QSP model of the immune response to SARS-CoV-2 for community development.
\newblock \emph{CPT: pharmacometrics \& systems pharmacology}, 10(1): 18--29.

\bibitem[{Dupont, Doucet, and Teh(2019)}]{dupont2019augmented}
Dupont, E.; Doucet, A.; and Teh, Y.~W. 2019.
\newblock Augmented neural odes.
\newblock In \emph{Advances in neural information processing systems}, 3140--3150.

\bibitem[{Feuerriegel et~al.(2024)Feuerriegel, Frauen, Melnychuk, Schweisthal, Hess, Curth, Bauer, Kilbertus, Kohane, and van~der Schaar}]{feuerriegel2024causal}
Feuerriegel, S.; Frauen, D.; Melnychuk, V.; Schweisthal, J.; Hess, K.; Curth, A.; Bauer, S.; Kilbertus, N.; Kohane, I.~S.; and van~der Schaar, M. 2024.
\newblock Causal machine learning for predicting treatment outcomes.
\newblock \emph{Nature Medicine}, 30(4): 958--968.

\bibitem[{Gandhi and Havlir(2021)}]{gandhi2021first}
Gandhi, M.; and Havlir, D.~V. 2021.
\newblock The First 100 Days of Severe Acute Respiratory Syndrome Coronavirus 2 (SARS-CoV-2) Control and Prevention in the United States.
\newblock \emph{Clinical Infectious Diseases}.
\newblock Proportion.

\bibitem[{Hess et~al.(2023)Hess, Melnychuk, Frauen, and Feuerriegel}]{hess2023bayesian}
Hess, K.; Melnychuk, V.; Frauen, D.; and Feuerriegel, S. 2023.
\newblock Bayesian neural controlled differential equations for treatment effect estimation.
\newblock \emph{arXiv preprint arXiv:2310.17463}.

\bibitem[{Hethcote(2000)}]{hethcote_mathematics_2000}
Hethcote, H.~W. 2000.
\newblock The {Mathematics} of {Infectious} {Diseases}.
\newblock \emph{SIAM Review}, 42(4): 599--653.

\bibitem[{Ho, Jain, and Abbeel(2020)}]{ho2020denoising}
Ho, J.; Jain, A.; and Abbeel, P. 2020.
\newblock Denoising diffusion probabilistic models.
\newblock \emph{Advances in neural information processing systems}, 33: 6840--6851.

\bibitem[{Holmdahl and Buckee(2020)}]{holmdahl2020wrong}
Holmdahl, I.; and Buckee, C. 2020.
\newblock Wrong but useful—what covid-19 epidemiologic models can and cannot tell us.
\newblock \emph{New England Journal of Medicine}, 383(4): 303--305.

\bibitem[{Horby et~al.(2021)Horby, Lim, Emberson, Mafham, Bell, Linsell, Staplin, Brightling, Ustianowski, Elmahi, and et~al.}]{Horby2021Dexamethasone}
Horby, P.; Lim, W.~S.; Emberson, J.~R.; Mafham, M.; Bell, J.~L.; Linsell, L.; Staplin, N.; Brightling, C.; Ustianowski, A.; Elmahi, E.; and et~al. 2021.
\newblock Dexamethasone in Hospitalized Patients with Covid-19.
\newblock \emph{The New England Journal of Medicine}, 384(8): 693--704.

\bibitem[{Huang et~al.(2024)Huang, Yang, Wang, and Park}]{huang2024diffusionpde}
Huang, J.; Yang, G.; Wang, Z.; and Park, J.~J. 2024.
\newblock DiffusionPDE: Generative PDE-solving under partial observation.
\newblock \emph{Advances in Neural Information Processing Systems}, 37: 130291--130323.

\bibitem[{Imbens(2004)}]{imbens2004nonparametric}
Imbens, G.~W. 2004.
\newblock Nonparametric estimation of average treatment effects under exogeneity: A review.
\newblock \emph{Review of Economics and statistics}, 86(1): 4--29.

\bibitem[{Jeanneret, Simon, and Jurie(2022)}]{jeanneret2022diffusion}
Jeanneret, G.; Simon, L.; and Jurie, F. 2022.
\newblock Diffusion models for counterfactual explanations.
\newblock In \emph{Proceedings of the Asian Conference on Computer Vision}, 858--876.

\bibitem[{Jhin et~al.(2021)Jhin, Jo, Kong, Jeon, and Park}]{jhin2021ace}
Jhin, S.~Y.; Jo, M.; Kong, T.; Jeon, J.; and Park, N. 2021.
\newblock Ace-node: Attentive co-evolving neural ordinary differential equations.
\newblock In \emph{Proceedings of the 27th ACM SIGKDD Conference on Knowledge Discovery \& Data Mining}, 736--745.

\bibitem[{Kain et~al.(2021)Kain, Childs, Becker, and Mordecai}]{kain2021chopping}
Kain, M.~P.; Childs, M.~L.; Becker, A.~D.; and Mordecai, E.~A. 2021.
\newblock Chopping the tail: how preventing superspreading can help to maintain COVID-19 control.
\newblock \emph{Epidemics}, 34: 100430.

\bibitem[{Kamarthi et~al.(2022)Kamarthi, Kong, Rodr{\'\i}guez, Zhang, and Prakash}]{kamarthi2022camul}
Kamarthi, H.; Kong, L.; Rodr{\'\i}guez, A.; Zhang, C.; and Prakash, B.~A. 2022.
\newblock CAMul: calibrated and accurate multi-view time-series forecasting.
\newblock In \emph{Proceedings of the ACM Web Conference 2022}, 3174--3185.

\bibitem[{Karniadakis et~al.(2021)Karniadakis, Kevrekidis, Lu, Perdikaris, Wang, and Yang}]{karniadakis2021physics}
Karniadakis, G.~E.; Kevrekidis, I.~G.; Lu, L.; Perdikaris, P.; Wang, S.; and Yang, L. 2021.
\newblock Physics-informed machine learning.
\newblock \emph{Nature Reviews Physics}, 3(6): 422--440.

\bibitem[{Kennedy(2023)}]{kennedy2023towards}
Kennedy, E.~H. 2023.
\newblock Towards optimal doubly robust estimation of heterogeneous causal effects.
\newblock \emph{Electronic Journal of Statistics}, 17(2): 3008--3049.

\bibitem[{Kennedy, Balakrishnan, and Wasserman(2023)}]{kennedy2023semiparametric}
Kennedy, E.~H.; Balakrishnan, S.; and Wasserman, L. 2023.
\newblock Semiparametric counterfactual density estimation.
\newblock \emph{Biometrika}, 110(4): 875--896.

\bibitem[{Khemakhem et~al.(2021)Khemakhem, Monti, Leech, and Hyvarinen}]{khemakhem2021causal}
Khemakhem, I.; Monti, R.; Leech, R.; and Hyvarinen, A. 2021.
\newblock Causal autoregressive flows.
\newblock In \emph{International conference on artificial intelligence and statistics}, 3520--3528. PMLR.

\bibitem[{Kidger et~al.(2020)Kidger, Morrill, Foster, and Lyons}]{kidger2020neural}
Kidger, P.; Morrill, J.; Foster, J.; and Lyons, T. 2020.
\newblock Neural controlled differential equations for irregular time series.
\newblock \emph{Advances in Neural Information Processing Systems}, 33: 6696--6707.

\bibitem[{Kim, Kim, and Kennedy(2018)}]{kim2018causal}
Kim, K.; Kim, J.; and Kennedy, E.~H. 2018.
\newblock Causal effects based on distributional distances.
\newblock \emph{arXiv preprint arXiv:1806.02935}.

\bibitem[{K{\"u}nzel et~al.(2019)K{\"u}nzel, Sekhon, Bickel, and Yu}]{kunzel2019metalearners}
K{\"u}nzel, S.~R.; Sekhon, J.~S.; Bickel, P.~J.; and Yu, B. 2019.
\newblock Metalearners for estimating heterogeneous treatment effects using machine learning.
\newblock \emph{Proceedings of the national academy of sciences}, 116(10): 4156--4165.

\bibitem[{Leon et~al.(2023)Leon, Tokarev, Bouchnita, and Volpert}]{leon2023modelling}
Leon, C.; Tokarev, A.; Bouchnita, A.; and Volpert, V. 2023.
\newblock Modelling of the Innate and Adaptive Immune Response to SARS Viral Infection, Cytokine Storm and Vaccination.
\newblock \emph{Vaccines (Basel)}, 11(1): 127.

\bibitem[{Li et~al.(2021)Li, Hu, Lu, Utsumi, Chakraborty, Sow, Madan, Li, Ghalwash, Shahn et~al.}]{li2021g}
Li, R.; Hu, S.; Lu, M.; Utsumi, Y.; Chakraborty, P.; Sow, D.~M.; Madan, P.; Li, J.; Ghalwash, M.; Shahn, Z.; et~al. 2021.
\newblock G-net: a recurrent network approach to g-computation for counterfactual prediction under a dynamic treatment regime.
\newblock In \emph{Machine Learning for Health}, 282--299. PMLR.

\bibitem[{Li and Rodr{\'\i}guez(2025)}]{li2025neural}
Li, R.; and Rodr{\'\i}guez, A. 2025.
\newblock Neural Conformal Control for Time Series Forecasting.
\newblock In \emph{Proceedings of the AAAI Conference on Artificial Intelligence}, volume~39, 18439--18447.

\bibitem[{Li et~al.(2020)Li, Lin, Wang, and Li}]{li2020scalable}
Li, X.; Lin, J.; Wang, Y.; and Li, Z. 2020.
\newblock Scalable neural stochastic differential equations for uncertainty quantification.
\newblock \emph{Advances in Neural Information Processing Systems}, 33.

\bibitem[{Li(2023)}]{li2023ts}
Li, Y. 2023.
\newblock Ts-diffusion: Generating highly complex time series with diffusion models.
\newblock \emph{arXiv preprint arXiv:2311.03303}.

\bibitem[{Lim(2018)}]{lim2018forecasting}
Lim, B. 2018.
\newblock Forecasting treatment responses over time using recurrent marginal structural networks.
\newblock \emph{Advances in neural information processing systems}, 31.

\bibitem[{Lin et~al.(2024)Lin, Shi, Han, and Gao}]{lin2024specstg}
Lin, L.; Shi, D.; Han, A.; and Gao, J. 2024.
\newblock Specstg: A fast spectral diffusion framework for probabilistic spatio-temporal traffic forecasting.
\newblock \emph{arXiv preprint arXiv:2401.08119}.

\bibitem[{Lyu and Wehby(2020)}]{lyu2020community}
Lyu, W.; and Wehby, G.~L. 2020.
\newblock Community Use Of Face Masks And COVID-19: Evidence From A Natural Experiment Of State Mandates In The US.
\newblock \emph{Health Affairs}, 39(8): 1419--1425.
\newblock Mask and alpha, delta.

\bibitem[{Ma et~al.(2024)Ma, Melnychuk, Schweisthal, and Feuerriegel}]{ma2024diffpo}
Ma, Y.; Melnychuk, V.; Schweisthal, J.; and Feuerriegel, S. 2024.
\newblock DiffPO: A causal diffusion model for learning distributions of potential outcomes.
\newblock In \emph{The Thirty-eighth Annual Conference on Neural Information Processing Systems}.

\bibitem[{Melnychuk, Frauen, and Feuerriegel(2022)}]{melnychuk2022causal}
Melnychuk, V.; Frauen, D.; and Feuerriegel, S. 2022.
\newblock Causal transformer for estimating counterfactual outcomes.
\newblock In \emph{International Conference on Machine Learning}, 15293--15329. PMLR.

\bibitem[{Melnychuk, Frauen, and Feuerriegel(2023)}]{melnychuk2023normalizing}
Melnychuk, V.; Frauen, D.; and Feuerriegel, S. 2023.
\newblock Normalizing flows for interventional density estimation.
\newblock In \emph{International Conference on Machine Learning}, 24361--24397. PMLR.

\bibitem[{Metcalf, Morris, and Park(2020)}]{metcalf2020mathematical}
Metcalf, C. J.~E.; Morris, D.~H.; and Park, S.~W. 2020.
\newblock Mathematical models to guide pandemic response.
\newblock \emph{Science}, 369(6502): 368--369.

\bibitem[{Mooij, Janzing, and Sch{\"o}lkopf(2013)}]{mooij2013ordinary}
Mooij, J.~M.; Janzing, D.; and Sch{\"o}lkopf, B. 2013.
\newblock From ordinary differential equations to structural causal models: the deterministic case.
\newblock In \emph{Proceedings of the Twenty-Ninth Conference on Uncertainty in Artificial Intelligence}, 440--448.

\bibitem[{Nag et~al.(2023)Nag, Zhu, Deng, Song, and Xiang}]{nag2023difftad}
Nag, S.; Zhu, X.; Deng, J.; Song, Y.-Z.; and Xiang, T. 2023.
\newblock Difftad: Temporal action detection with proposal denoising diffusion.
\newblock In \emph{Proceedings of the IEEE/CVF International Conference on Computer Vision}, 10362--10374.

\bibitem[{Nguyen and et~al.(2021)}]{nguyen2021impact}
Nguyen, V.~D.; and et~al. 2021.
\newblock Impact of State Mask-Wearing Requirement on the Transmission of COVID-19 in the United States.
\newblock \emph{Innovation}, 2(4): 100176.
\newblock Mask.

\bibitem[{Nie and Wager(2021)}]{nie2021quasi}
Nie, X.; and Wager, S. 2021.
\newblock Quasi-oracle estimation of heterogeneous treatment effects.
\newblock \emph{Biometrika}, 108(2): 299--319.

\bibitem[{Qian, Alaa, and van~der Schaar(2020)}]{qian2020and}
Qian, Z.; Alaa, A.~M.; and van~der Schaar, M. 2020.
\newblock When and how to lift the lockdown? global covid-19 scenario analysis and policy assessment using compartmental gaussian processes.
\newblock \emph{Advances in Neural Information Processing Systems}, 33: 10729--10740.

\bibitem[{Qian et~al.(2021)Qian, Zame, Fleuren, Elbers, and van~der Schaar}]{qian2021integrating}
Qian, Z.; Zame, W.; Fleuren, L.; Elbers, P.; and van~der Schaar, M. 2021.
\newblock Integrating expert ODEs into neural ODEs: pharmacology and disease progression.
\newblock \emph{Advances in Neural Information Processing Systems}, 34: 11364--11383.

\bibitem[{Reynaud et~al.(2023)Reynaud, Qiao, Dombrowski, Day, Razavi, Gomez, Leeson, and Kainz}]{reynaud2023feature}
Reynaud, H.; Qiao, M.; Dombrowski, M.; Day, T.; Razavi, R.; Gomez, A.; Leeson, P.; and Kainz, B. 2023.
\newblock Feature-conditioned cascaded video diffusion models for precise echocardiogram synthesis.
\newblock In \emph{International Conference on Medical Image Computing and Computer-Assisted Intervention}, 142--152. Springer.

\bibitem[{Reynaud et~al.(2022)Reynaud, Vlontzos, Dombrowski, Gilligan~Lee, Beqiri, Leeson, and Kainz}]{reynaud2022d}
Reynaud, H.; Vlontzos, A.; Dombrowski, M.; Gilligan~Lee, C.; Beqiri, A.; Leeson, P.; and Kainz, B. 2022.
\newblock D’artagnan: Counterfactual video generation.
\newblock In \emph{International Conference on Medical Image Computing and Computer-Assisted Intervention}, 599--609. Springer.

\bibitem[{Robins(1986)}]{robins1986new}
Robins, J. 1986.
\newblock A new approach to causal inference in mortality studies with a sustained exposure period—application to control of the healthy worker survivor effect.
\newblock \emph{Mathematical modelling}, 7(9-12): 1393--1512.

\bibitem[{Robins(1994)}]{robins1994correcting}
Robins, J.~M. 1994.
\newblock Correcting for non-compliance in randomized trials using structural nested mean models.
\newblock \emph{Communications in Statistics-Theory and methods}, 23(8): 2379--2412.

\bibitem[{Robins, Hernan, and Brumback(2000)}]{robins2000marginal}
Robins, J.~M.; Hernan, M.~A.; and Brumback, B. 2000.
\newblock Marginal structural models and causal inference in epidemiology.

\bibitem[{Rodr{\'\i}guez et~al.(2023)Rodr{\'\i}guez, Cui, Ramakrishnan, Adhikari, and Prakash}]{rodriguez2023einns}
Rodr{\'\i}guez, A.; Cui, J.; Ramakrishnan, N.; Adhikari, B.; and Prakash, B.~A. 2023.
\newblock Einns: epidemiologically-informed neural networks.
\newblock In \emph{Proceedings of the AAAI conference on artificial intelligence}, volume~37, 14453--14460.

\bibitem[{Rodr{\'\i}guez et~al.(2024)Rodr{\'\i}guez, Kamarthi, Agarwal, Ho, Patel, Sapre, and Prakash}]{rodriguez2024machine}
Rodr{\'\i}guez, A.; Kamarthi, H.; Agarwal, P.; Ho, J.; Patel, M.; Sapre, S.; and Prakash, B.~A. 2024.
\newblock Machine learning for data-centric epidemic forecasting.
\newblock \emph{Nature Machine Intelligence}, 6(10): 1122--1131.

\bibitem[{Rombach et~al.(2022)Rombach, Blattmann, Lorenz, Esser, and Ommer}]{rombach2022high}
Rombach, R.; Blattmann, A.; Lorenz, D.; Esser, P.; and Ommer, B. 2022.
\newblock High-resolution image synthesis with latent diffusion models.
\newblock In \emph{Proceedings of the IEEE/CVF conference on computer vision and pattern recognition}, 10684--10695.

\bibitem[{Rubin(1978)}]{rubin1978bayesian}
Rubin, D.~B. 1978.
\newblock Bayesian inference for causal effects: The role of randomization.
\newblock \emph{The Annals of statistics}, 34--58.

\bibitem[{Rubin(2005)}]{rubin2005causal}
Rubin, D.~B. 2005.
\newblock Causal inference using potential outcomes: Design, modeling, decisions.
\newblock \emph{Journal of the American Statistical Association}, 100(469): 322--331.

\bibitem[{Salsa(2015)}]{salsa2015partial}
Salsa, S. 2015.
\newblock \emph{Partial differential equations in action}, volume~1.
\newblock Springer.

\bibitem[{Sanchez and Tsaftaris(2022)}]{sanchez2022diffusion}
Sanchez, P.; and Tsaftaris, S.~A. 2022.
\newblock Diffusion causal models for counterfactual estimation.
\newblock \emph{arXiv preprint arXiv:2202.10166}.

\bibitem[{Sanchez-Martin, Rateike, and Valera(2021)}]{sanchez2021vaca}
Sanchez-Martin, P.; Rateike, M.; and Valera, I. 2021.
\newblock Vaca: Design of variational graph autoencoders for interventional and counterfactual queries.
\newblock \emph{arXiv preprint arXiv:2110.14690}.

\bibitem[{Sch{\"o}lkopf(2022)}]{scholkopf2022causality}
Sch{\"o}lkopf, B. 2022.
\newblock Causality for machine learning.
\newblock In \emph{Probabilistic and causal inference: The works of Judea Pearl}, 765--804.

\bibitem[{Seedat et~al.(2022)Seedat, Imrie, Bellot, Qian, and van~der Schaar}]{seedat2022continuous}
Seedat, N.; Imrie, F.; Bellot, A.; Qian, Z.; and van~der Schaar, M. 2022.
\newblock Continuous-time modeling of counterfactual outcomes using neural controlled differential equations.
\newblock \emph{arXiv preprint arXiv:2206.08311}.

\bibitem[{Shalit, Johansson, and Sontag(2017)}]{shalit2017estimating}
Shalit, U.; Johansson, F.~D.; and Sontag, D. 2017.
\newblock Estimating individual treatment effect: generalization bounds and algorithms.
\newblock In \emph{International conference on machine learning}, 3076--3085. PMLR.

\bibitem[{Shao et~al.(2024)Shao, Feng, Lu, Zhang, and Zou}]{shao2024data}
Shao, P.; Feng, J.; Lu, J.; Zhang, P.; and Zou, C. 2024.
\newblock Data-driven and knowledge-guided denoising diffusion model for flood forecasting.
\newblock \emph{Expert Systems with Applications}, 244: 122908.

\bibitem[{Shi et~al.(2019)Shi, Chen, Yu et~al.}]{shi2019ganite}
Shi, Z.; Chen, Z.; Yu, Y.; et~al. 2019.
\newblock GANITE: Interpretable Counterfactual Learning with Generative Models.
\newblock \emph{Proceedings of the 36th International Conference on Machine Learning}.

\bibitem[{Sohl-Dickstein et~al.(2015)Sohl-Dickstein, Weiss, Maheswaranathan, and Ganguli}]{sohl2015deep}
Sohl-Dickstein, J.; Weiss, E.; Maheswaranathan, N.; and Ganguli, S. 2015.
\newblock Deep unsupervised learning using nonequilibrium thermodynamics.
\newblock In \emph{International conference on machine learning}, 2256--2265. PMLR.

\bibitem[{Sohn, Lee, and Yan(2015)}]{sohn2015learning}
Sohn, K.; Lee, H.; and Yan, X. 2015.
\newblock Learning structured output representation using deep conditional generative models.
\newblock \emph{Advances in neural information processing systems}, 28.

\bibitem[{Song and Ermon(2019)}]{song2019generative}
Song, Y.; and Ermon, S. 2019.
\newblock Generative modeling by estimating gradients of the data distribution.
\newblock \emph{Advances in neural information processing systems}, 32.

\bibitem[{Team(2021)}]{ihme2021modeling}
Team, I. C.-.~F. 2021.
\newblock Modeling COVID-19 scenarios for the United States.
\newblock \emph{Nature medicine}, 27(1): 94--105.

\bibitem[{{U.S. Census Bureau}(2020)}]{uscensus2020}
{U.S. Census Bureau}. 2020.
\newblock City and Town Population Totals: 2020-2022.
\newblock Available at: \url{https://www.census.gov/data/tables/time-series/demo/popest/2020s-total-cities-and-towns.html}.
\newblock Accessed: Jan 2025.

\bibitem[{Wang et~al.(2018)Wang, Zhou, Song, and Sherwood}]{wang2018quantile}
Wang, L.; Zhou, Y.; Song, R.; and Sherwood, B. 2018.
\newblock Quantile-optimal treatment regimes.
\newblock \emph{Journal of the American Statistical Association}, 113(523): 1243--1254.

\bibitem[{Wang et~al.(2023)Wang, Xu, Feng, Lin, He, and Chua}]{wang2023diffusion}
Wang, W.; Xu, Y.; Feng, F.; Lin, X.; He, X.; and Chua, T.-S. 2023.
\newblock Diffusion recommender model.
\newblock In \emph{Proceedings of the 46th International ACM SIGIR Conference on Research and Development in Information Retrieval}, 832--841.

\bibitem[{Wu, Leung, and Leung(2020)}]{wu2020nowcasting}
Wu, J.~T.; Leung, K.; and Leung, G.~M. 2020.
\newblock Nowcasting and forecasting the potential domestic and international spread of the 2019-nCoV outbreak originating in Wuhan, China: a modelling study.
\newblock \emph{The lancet}, 395(10225): 689--697.

\bibitem[{Wu et~al.(2024)Wu, Zhou, Chen, and Zhu}]{wu2024counterfactual}
Wu, S.; Zhou, W.; Chen, M.; and Zhu, S. 2024.
\newblock Counterfactual generative models for time-varying treatments.
\newblock In \emph{Proceedings of the 30th ACM SIGKDD Conference on Knowledge Discovery and Data Mining}, 3402--3413.

\bibitem[{Wyatt et~al.(2022)Wyatt, Leach, Schmon, and Willcocks}]{wyatt2022anoddpm}
Wyatt, J.; Leach, A.; Schmon, S.~M.; and Willcocks, C.~G. 2022.
\newblock Anoddpm: Anomaly detection with denoising diffusion probabilistic models using simplex noise.
\newblock In \emph{Proceedings of the IEEE/CVF Conference on Computer Vision and Pattern Recognition}, 650--656.

\bibitem[{Xu and Xie(2021)}]{xu2021conformal}
Xu, C.; and Xie, Y. 2021.
\newblock Conformal prediction interval for dynamic time-series.
\newblock In \emph{International Conference on Machine Learning}, 11559--11569. PMLR.

\bibitem[{Yin et~al.(2021)Yin, Le~Guen, Dona, de~B{\'e}zenac, Ayed, Thome, and Gallinari}]{yin2021augmenting}
Yin, Y.; Le~Guen, V.; Dona, J.; de~B{\'e}zenac, E.; Ayed, I.; Thome, N.; and Gallinari, P. 2021.
\newblock Augmenting physical models with deep networks for complex dynamics forecasting.
\newblock \emph{Journal of Statistical Mechanics: Theory and Experiment}, 2021(12): 124012.

\bibitem[{Yoon, Jordon, and Van Der~Schaar(2018)}]{yoon2018ganite}
Yoon, J.; Jordon, J.; and Van Der~Schaar, M. 2018.
\newblock GANITE: Estimation of individualized treatment effects using generative adversarial nets.
\newblock In \emph{International conference on learning representations}.

\bibitem[{Yuan and Qiao(2024)}]{yuan2024diffusion}
Yuan, X.; and Qiao, Y. 2024.
\newblock Diffusion-TS: Interpretable Diffusion for General Time Series Generation.
\newblock In \emph{The Twelfth International Conference on Learning Representations}.

\bibitem[{Zhang, Liu, and Li(2021)}]{zhang2021treatment}
Zhang, W.; Liu, L.; and Li, J. 2021.
\newblock Treatment effect estimation with disentangled latent factors.
\newblock In \emph{Proceedings of the AAAI Conference on Artificial Intelligence}, volume~35, 10923--10930.

\end{thebibliography}
